\documentclass[peerreview, onecolumn]{IEEEtran}
\usepackage{cite}
\usepackage{url}
\usepackage[utf8]{inputenc}
\usepackage{booktabs}
\usepackage{graphicx}
\usepackage{amsmath,amssymb}
\usepackage{hyperref}
\usepackage{svg}
\usepackage{multirow}
\usepackage{float}
\usepackage[table]{xcolor}
\usepackage{algorithm}
\usepackage{algorithmic}
\usepackage{array}
\usepackage{tabularx}
\usepackage{subcaption}
\usepackage{listings}
\usepackage{xcolor}
\usepackage{newunicodechar}
\usepackage{fancyhdr}
\usepackage{titling}

\newunicodechar{∼}{\textasciitilde}
\newunicodechar{−}{\textminus}
\lstdefinestyle{jsonl}{
    basicstyle=\ttfamily\small,          
    breaklines=true,                     
    breakatwhitespace=false,
    showstringspaces=false,
    numbers=left,                        
    numberstyle=\tiny\color{gray},
    frame=single,                        
    literate=
        {á}{{\'a}}1  {é}{{\'e}}1  {í}{{\'i}}1  {ó}{{\'o}}1  {ú}{{\'u}}1  
        {ä}{{\"a}}1  {ë}{{\"e}}1  {ï}{{\"i}}1  {ö}{{\"o}}1  {ü}{{\"u}}1
        {ñ}{{\~n}}1
        {_}{{\textunderscore}}1          
        {\{}{{\textbraceleft}}1          
        {\}}{{\textbraceright}}1,
    escapechar=@,                        
}

\begin{document}


\title{NebulaExp-8B: An Empirical Post-Training Pipeline via Full-Scale Ablation Research}

\author{ZTE NebulaL0 Post-Training Team \\
}
\date{\today}

\maketitle
\thispagestyle{plain}


\begin{abstract}
Post-training alignment determines the reasoning and human preference following capabilities of large language models, yet most existing works withhold detailed data construction, filtering rules and training recipes, which hinders community reproducibility and lightweight model optimization. This work presents NebulaExp, a fully transparent, ablation-driven post-training pipeline built on Qwen3-8B-base, covering two orthogonal model branches: general instruct model and complex reasoning-specialized model. We curate a raw corpus of 3.84M multi-source SFT samples and a 200K verifiable RL candidate pool, and design an end-to-end data processing stack including response distillation, multi-dimensional cross-verification filtering, fine-grained difficulty grading, task classification and diversity-aware sampling. Controlled ablation experiments are conducted over SFT, GRPO reinforcement learning (RL), and On-Policy Distillation (OPD) modules to quantify the impact of data quality, domain mixture ratio, sample difficulty and teacher selection. For the Instruct branch, our three-stage optimized supervised fine-tuning approach NebulaExp-Ins-SFT improves the average benchmark score from the 55.01 baseline of Qwen3-8B-nothink to 60.99. GRPO reinforcement learning then further elevates the average score to 61.85. We empirically identify two key trade-offs: mathematical reasoning and code generation require opposite data distribution preferences, and long-form general text delivers cross-transfer gains for formal logical reasoning. For the Reasoning branch, medium-difficulty GRPO RL improves average reasoning score from 73.88 to 75.17. To address RL’s dependency on task verifiers, we systematically investigate single-teacher and multi-teacher OPD (MOPD): utilizing merely 4K instruction-following samples and outperforms RL baseline by 3.26 points on IFEval with +4.43 average overall gain; MOPD fuses four domain-specialist teachers with merely 10K samples, lifting average performance by 4.18 over the base model. Notably, cross-teacher knowledge recombination enables the student model to exceed the upper performance bound of individual teachers on mathematical reasoning benchmarks. Our core findings prove that data correctness filtering acts as the first-order optimization factor, while teacher distribution compatibility outweighs model scale for distillation. This report provides a fully reproducible empirical post-training recipe for 8B-scale LLMs, and comprehensively dissects the capability trade-offs among instruction adherence, mathematical reasoning, code generation and general knowledge.
\end{abstract}

\section{Introduction}
Post-training is a critical phase for large language models after pre-training, focusing on human-value alignment, usability improvement, and safety enhancement. In recent frontier model families, such as GPT~\cite{singh2025openai}, GLM~\cite{glm}, DeepSeek~\cite{deepseek_v4}, and Llama~\cite{llama}, the increasing emphasis on alignment data, reasoning data, reinforcement learning, and safety evaluation shows that future competition among large models will depend not only on pre-training scale, but also on corpus quality and the efficiency of the post-training feedback loop. Therefore, we focus on the post-training stage and organize the discussion around three closely related components: evaluation, corpus construction, and training methodology.

It is critical to select high-quality evaluation benchmarks at the initial stage of model training. A reliable evaluation framework is also essential, as model competence manifests across diverse dimensions rather than a single metric. Accordingly, evaluations commonly draw upon multiple publicly available and authoritative benchmarks spanning instruction following, code generation, mathematical reasoning, scientific problem-solving, commonsense reasoning and logical deduction. Early studies extensively employed classic benchmarks including MMLU~\cite{hendrycks2020measuring}, GSM8K~\cite{cobbe2021training} and HumanEval~\cite{chen2021evaluating}. As model capabilities keep advancing, evaluation paradigms have gradually shifted toward more difficult, domain-specialized benchmarks. Consequently, recent evaluations increasingly incorporate datasets such as IFEVal~\cite{ifeval}, LiveCodeBench~\cite{livecodebench}, Math-500~\cite{math500}, AIME'25~\cite{aime2025}, GPQA-Diamond~\cite{rein2023gpqa}, MMLU-Redux~\cite{mmu_redux}, AutoLogi~\cite{autologi} and ZebraLogic~\cite{zebralogic}.

While rigorous evaluation benchmarks enable accurate performance quantification, the upper bound of model performance is fundamentally determined by the quality of training data. Corpus construction is an essential part of post-training, because the model can only learn behaviors that are represented in its training corpus and, when possible, supported by verifiable supervision. Due to the limited availability and inconsistent quality of open-source corpora, corpus construction has gradually evolved into a process centered on synthetic data generation, correctness verification, and reasoning alignment. In this process, strong teacher models, such as DeepSeek-V4~\cite{deepseek_v4}, Qwen3.5~\cite{qwen3}, and GPT-5~\cite{singh2025openai}, are often used to synthesize high-quality samples. However, low-quality cases may still occur after data generation. Therefore, raw data must undergo a series of quality verification steps, including rule-based filtering, model-based scoring, and task-specific verification such as mathematical answer checking or code execution tests~\cite{wang2025nemotron}. Lastly, to further improve data quality, reasoning alignment is applied to ensure that the model's internal logic and intermediate reasoning steps are consistent and interpretable, with their quality judged by another evaluation model. Each step is essential, and every detail in the process plays a direct role in determining the quality of the final corpus. Therefore, how to construct a data processing pipeline has attracted extensive research attention.

With the growing demand for both model capability and training efficiency, post-training paradigms have undergone continuous refinement. Early post-training pipelines such as Qwen1~\cite{bai2023qwen} commonly rely on Supervised Fine-Tuning (SFT) followed by reinforcement learning from human feedback built upon Proximal Policy Optimization (PPO). In this setting, SFT is used to establish basic instruction-following behavior, while reinforcement learning further optimizes model outputs with the help of a separately trained reward model. Although this paradigm proved effective, it also introduced considerable costs in human annotation, reward modeling, value estimation, and reinforcement learning optimization. To solve this problem, DeepSeek-V2~\cite{liu2024deepseek-v2} adopts GRPO after SFT, which does not require a value network and replaces absolute value estimation with group-wise relative comparison. In parallel, Qwen2~\cite{qwen22024tech} uses Direct Preference Optimization (DPO), which directly optimizes preference pairs without explicitly training a reward model. Then, Qwen2.5~\cite{qwen25_2024} further combines offline DPO with online GRPO, implying a general trend of shifting from static preference optimization to hybrid offline--online training strategies. As reasoning capability became an increasingly important objective, post-training methods further evolved from general alignment toward reasoning-oriented optimization. DeepSeek-V3~\cite{liu2024deepseek} continued the supervised fine-tuning and GRPO-based training strategy, while using reasoning distillation to support the development of stronger reasoning models. DeepSeek-R1~\cite{guo2025deepseek} proposed a more systematic post-training recipe, incorporating cold-start data, reasoning-oriented reinforcement learning, rejection sampling, and general reinforcement learning to elicit long-chain reasoning behaviors. More recent models have extended this direction through multi-stage reinforcement learning, outcome-guided preference optimization, agentic reinforcement learning, real-environment interaction, specialist model training, and on-policy distillation. These developments indicate that the choice of post-training paradigm and optimization method has become a critical factor in improving both model performance and training efficiency.

Despite rapid advances in corpus construction and post-training algorithms, the practical details of data processing and training methodologies that govern post-training performance remain insufficiently transparent. This lack of transparency makes it difficult for researchers to reference prior work and reproduce experiments.

To address this limitation, we introduce NebulaExp, a fully open post-training study built on Qwen3-8B-base~\cite{teamqwen3}. Rather than only reporting final model performance, NebulaExp provides a transparent account of the entire post-training process, including data sources, corpus construction, training methodology, stage-wise training design, intermediate results, ablation studies, empirical observations, and evaluation protocols. We first focus on data processing, which plays a central role in determining post-training effectiveness. NebulaExp presents the full corpus construction process, including data filtering, response-style distillation, difficulty grading, and diversity sampling. After establishing a complete data processing pipeline, we further conduct and report ablation experiments across different training stages, including supervised fine-tuning, reinforcement learning, and OPD optimization, together with the empirical observations obtained during these stages. Finally, under our evaluation framework, the trained NebulaExp models achieve stronger overall performance than the corresponding Qwen3-8B-nothink or Qwen3-8B-thinking baselines under their respective evaluation protocols. This result suggests that a carefully designed and transparent post-training pipeline can substantially improve model capability without modifying the base model architecture. Ultimately, the core contribution of this paper lies in revealing the internal mechanisms and empirical trade-offs of the post-training pipeline, rather than merely chasing absolute SOTA leaderboard rankings.

Specifically, the contributions of our work include:
\begin{itemize}
    \item We construct a complete data processing pipeline for post-training. The pipeline covers multi-source data collection, response-style distillation, rule-based and model-based filtering, task-specific correctness verification, difficulty grading, classification, and diversity-aware sampling. These components provide practical and reusable tools for constructing high-quality SFT corpora from open-source data.
    \item We conduct a series of ablation experiments based on Qwen3-8B-base. In particular, we study how data quality, sample difficulty, response length, curriculum design, and data-type mixture ratios affect downstream capability, and we provide a reference strategy for selecting and mixing SFT corpora. Based on this training recipe, our trained models achieve stronger overall performance than Qwen3-8B-nothink or Qwen3-8B-thinking under the corresponding evaluation setting.
    \item We perform Reinforcement Learning (RL) ablation experiments based on SFT models, including RL algorithm ablations for instruct models and GRPO-based medium-difficulty RL experiments for reasoning models. With verifiable reward formulation and multi-configuration RL tuning, we empirically derive a pragmatic RL recipe to boost model performance across reasoning, coding, and instruction following.
    \item We investigate On-Policy Distillation (OPD) as a verifier-free post-training paradigm, covering both single-teacher and multi-teacher OPD (MOPD) settings. With only 4K IF samples, single-teacher OPD surpasses the RL baseline on IFEval by up to 3.26 points and elevates the general capability average by up to 4.43 points. In the multi-teacher setting, MOPD fuses four domain-specialist teachers via only 10K samples into a unified model achieving a 4.18-point average improvement over the base model, with the student surpassing individual teacher boundaries on mathematical reasoning through cross-teacher knowledge recombination.
\end{itemize}

We organize the remainder of this report as follows. Section~\ref{sec:data_curation} details the corpus sources and preprocessing pipeline for data employed in the SFT and RL stages. Following this pipeline, we curate 3.84 million raw samples for SFT and construct a 200K verifiable multi-domain RL candidate pool, from which smaller filtered subsets are sampled for the actual RL experiments. Section~\ref{sec:eval_framework} establishes the evaluation framework, covering benchmark selection across general knowledge, alignment, mathematical reasoning, and coding, together with the unified evaluation protocol used throughout all experiments. Section~\ref{sec:instruct_training} presents ablation studies on the SFT and RL modules of the Instruct model and summarizes practical training insights from experimental outcomes. Section~\ref{sec:reasoning_training} presents ablation studies on the SFT, RL, and OPD modules of the Reasoning model, based on which we summarize our key findings. Finally, Section~\ref{sec:discussion} concludes with a summary of findings, broader discussion, and limitations of the current work.

\section{Data Curation}
\label{sec:data_curation}
\subsection{SFT Data Curation}
\begin{figure}[t]
\centering
\includegraphics[width=\columnwidth]{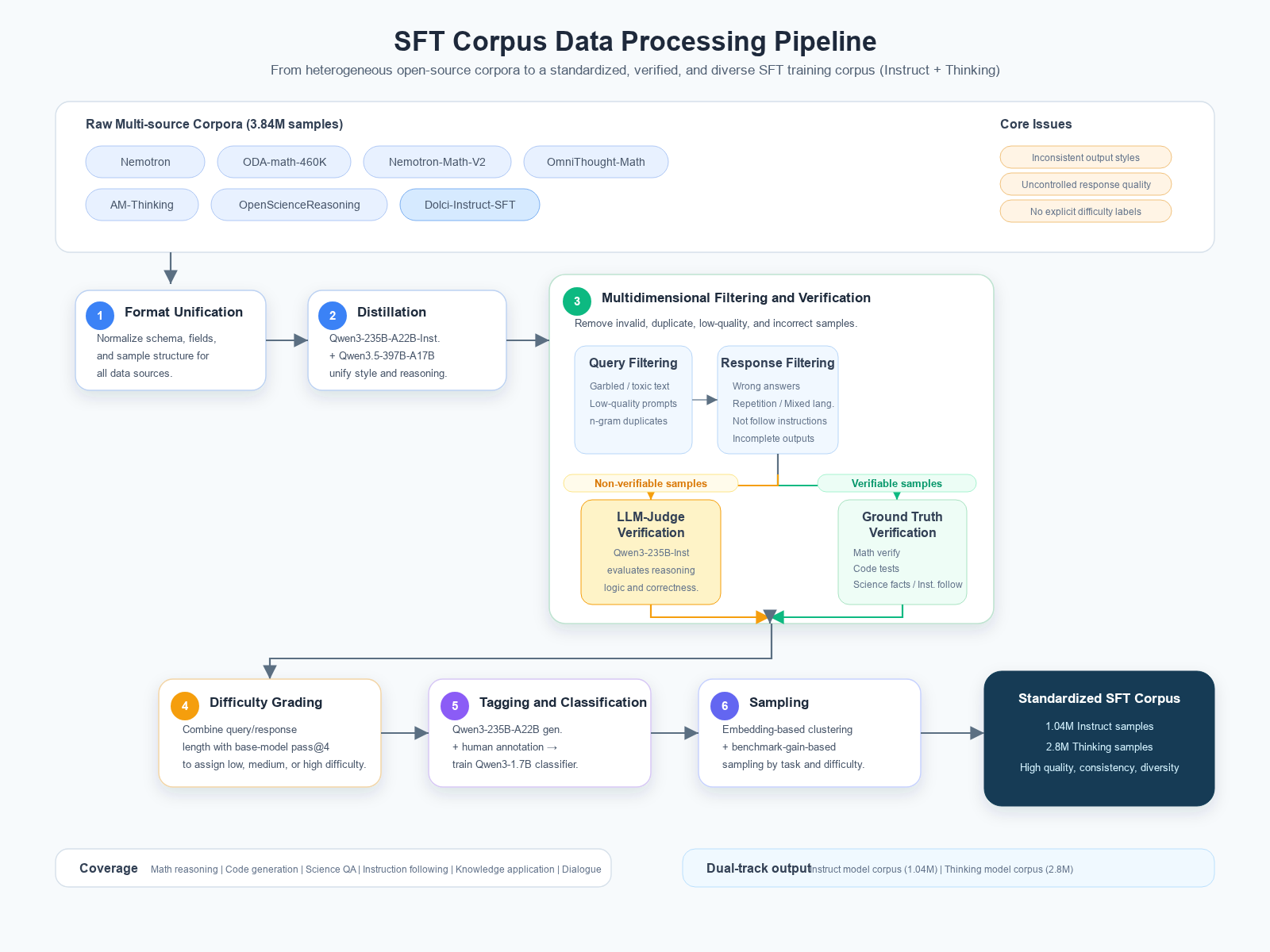}
\caption{Overview of the SFT data curation pipeline. Raw samples from public datasets undergo format unification, distillation, filtering, difficulty grading, classification, and diversity-oriented sampling to produce the final training corpus. Each stage addresses a specific data quality challenge identified in our analysis.}
\label{fig:SFT data pipeline}
\end{figure}

We curate a corpus of 3.84M samples from publicly available datasets, including Nemotron-cascade~\cite{wang2025nemotron}, ODA-math-460K~\cite{gao2025closing}, Nemotron-Math-V2~\cite{du2025nemotron}, OmniThought-Math~\cite{cai2025reasoning}, AM-Thinking~\cite{tian2025not}, OpenScienceReasoning, and Dolci-Instruct-SFT~\cite{olmo2025olmo3}. The corpus covers instruction following, math, code, knowledge application, general STEM problems, and daily dialogue.
\begin{figure}[htbp]
    \centering
    \begin{subfigure}[b]{0.48\textwidth}
        \centering
        \includegraphics[width=\linewidth]{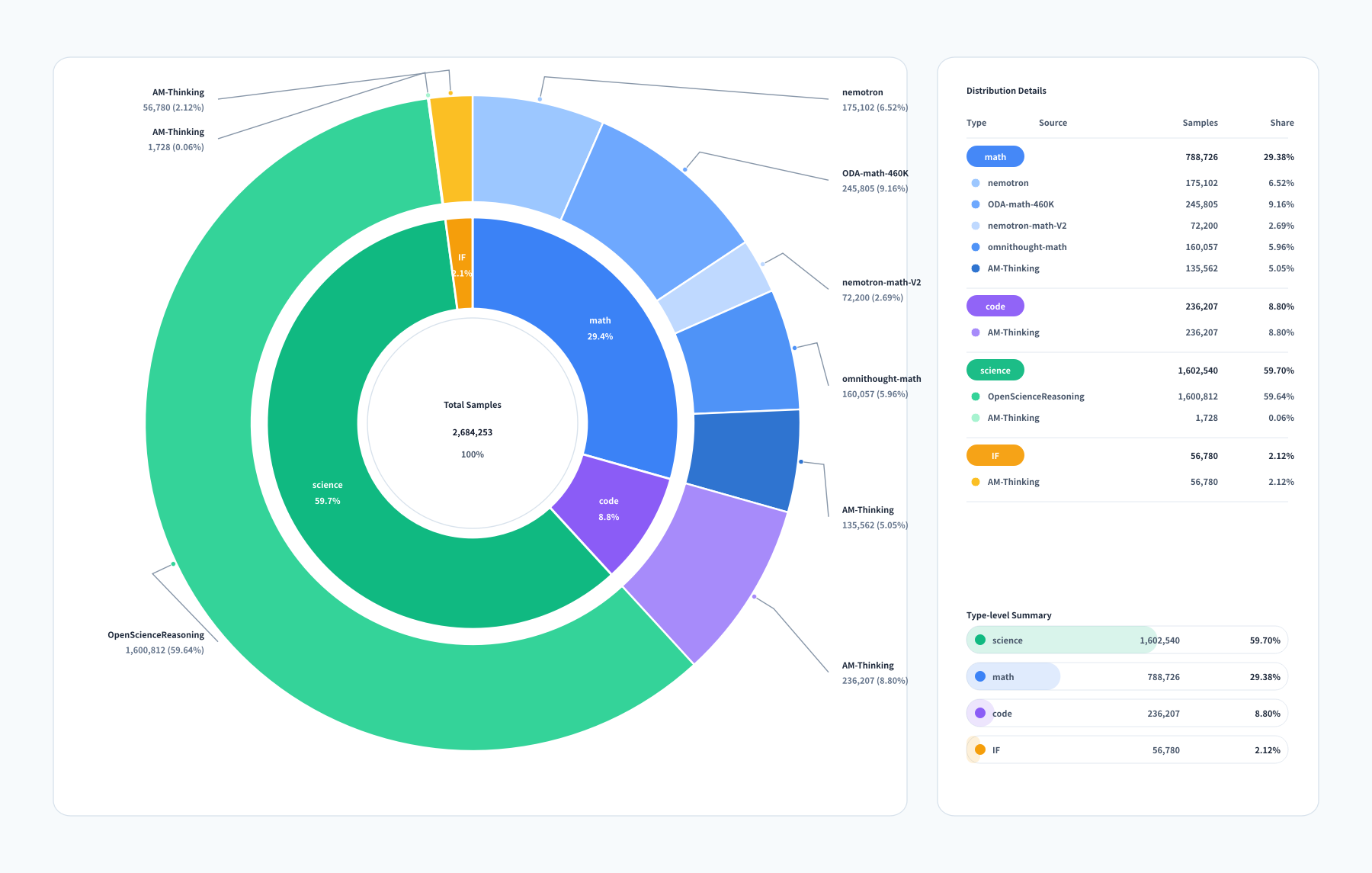}
        \caption{Thinking data distribution.}
        \label{fig:sub1}
    \end{subfigure}
    \hfill
    \begin{subfigure}[b]{0.48\textwidth}
        \centering
        \includegraphics[width=\linewidth]{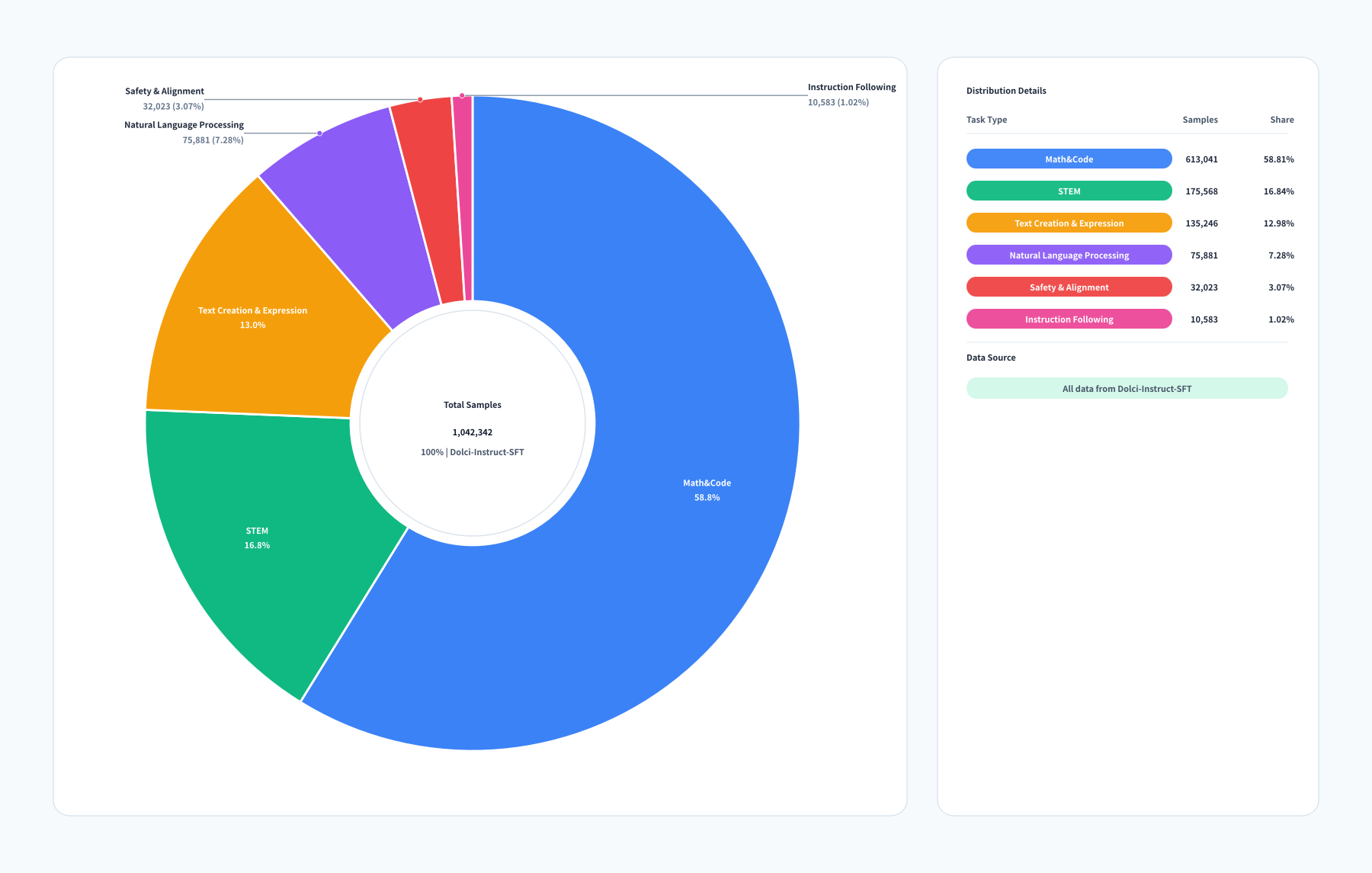}
        \caption{Instruct data distribution.}
        \label{fig:sub2}
    \end{subfigure}
    \caption{SFT data distribution, including data source and task type.}
    \label{fig:data_distribution}
\end{figure}
Collecting samples from publicly available datasets expands the number of training samples. However, directly combining these corpora introduces several challenges. The response styles are inconsistent: because the original data are distilled by different large models, the dialogue logic, response paradigms, language styles, and content structures vary substantially across samples. Response quality also differs widely: many publicly available datasets lack ground-truth answers and contain low-quality samples with incorrect answers, logical conflicts, or redundant content. Furthermore, the difficulty levels of the samples are undifferentiated---the datasets span a broad range of difficulty but use no shared, standardized grading rule. To address these issues, we design a data processing pipeline consisting of format unification, distillation, filtering, difficulty grading, classification and sampling. The detailed processing procedure and core methods are shown in Fig.~\ref{fig:SFT data pipeline} and described below.

To separately train the Instruct model and the reasoning model, we perform corresponding distillation and processing on corpora of both formats. After processing, we obtain 1.04M instruct samples and 2.8M thinking samples.

\subsubsection{Distillation}
To unify the mixed language styles and inconsistent response paradigms, we use Qwen3-235B-A22B-Instruct~\cite{yang2025qwen3} and Qwen3.5-397B-A17B~\cite{teamqwen3} to generate responses for each instruction. This step aligns the dialogue logic, reasoning patterns, language style, and content structure of all samples, standardizing the output format and improving the overall consistency of the SFT corpus. It also promotes more stable convergence during training and reduces bias caused by stylistic inconsistency.

\subsubsection{Filtering}
The public datasets exhibit wide variation in response quality and lack standardized ground truth. To handle these problems, we build a pipeline that combines rule-based filtering, model-score filtering, and task-specific ground-truth generation and verification. This pipeline identifies problematic samples from multiple perspectives and removes low-quality, invalid, duplicate, and incorrect data, improving the overall reliability of the corpus. Following the processing method of Qwen3~\cite{yang2025qwen3}, we adopt a bidirectional filtering approach with three phases: query filtering, response filtering, and answer verification.

\textbf{Query filtering:} The pipeline applies three sequential steps. It removes invalid queries through rules, targeting garbled text, abnormal characters, and toxic or harmful content. A model-based step using Qwen3-235B-A22B-Instruct~\cite{yang2025qwen3} then identifies and filters out low-quality queries with language confusion, missing instructions, or semantic repetition. An $n$-gram matching step removes duplicate queries to reduce redundancy.

\textbf{Response filtering:} The pipeline targets outputs that reduce training quality and reliability. It filters out six types of low-quality responses: those with incorrect final answers, highly repetitive content, mixed-language or stylistically disordered outputs, interrupted reasoning chains, responses that fail to follow instructions, and incomplete responses.

\textbf{Answer verification:} We use different verification strategies for verifiable and unverifiable samples. 

For samples with verifiable ground truth, we apply strict alignment verification and retain only those that meet task-specific correctness criteria.  
For mathematical samples, the LLM evaluates both the logical validity of the reasoning process and the accuracy of the final calculation, supplemented by tool-based checking in an isolated Docker environment and multi-model rejection sampling. 
For code data with test cases, a sample is retained only when all test cases are passed, including function-call tests, standard input-output tests, and assert tests. For code data without test cases, the LLM evaluates each generated solution according to a five-dimensional rubric, covering functionality and correctness, code readability and style, maintainability and architecture, testing and validation quality, and performance and security, with automatically generated test cases providing additional verification. 
For science data, verification is conducted according to question type: multiple-choice questions are checked through multi-model cross-validation, option perturbation tests, and reverse correctness derivation, while open-ended questions are verified by decomposing the answer into core knowledge points and assessing their factual consistency against objective evidence.
For instruction-following datasets, we filter responses to retain only those passing rule-based validation. We first parse instruction IDs and parameter lists from ground-truth annotations, then perform granular constraint-wise validation against model outputs via 54 pre-defined instruction checkers spanning keyword matching, length restriction, format compliance, case specification, and other verification rules to confirm full adherence to all given instructions~\cite{zhou2023instruction}.

For unverifiable samples without available ground truth, we employ an LLM-Judge verification mechanism: Qwen3-235B-A22B-Instruct~\cite{yang2025qwen3} evaluates each sample, and only those that pass its judgment are retained. The key prompt template is shown in Algorithm~\ref{alg:llm_judge_filter}.

\begin{algorithm}[b]
\caption{Judge Prompt Template}
\label{alg:llm_judge_filter}
\begin{algorithmic}[1]
\STATE \textbf{Role:} You are an expert evaluator with extremely high capability in mathematics, programming, reasoning, and instruction-following. Your task is to judge whether a reference Output is correct based on a given Instruction.
\STATE \textbf{Input Data:}
\STATE \quad 1. Instruction: \{instruction\}
\STATE \quad 2. Output (Reference): \{output\}

\STATE \textbf{Task Instructions:}
\STATE Please follow the logic below strictly:
\STATE \quad 1. \textbf{Accuracy Check}: Compare the Instruction and the Output. If they are consistent / matching / arrive at the same final conclusion, then judge it as "correct and the two are consistent".
\STATE \quad 2. \textbf{Consistency Re-verification}:  If the two are inconsistent, please re-answer, then re-examine whether the Output contains equivalent expressions, calculation errors, or wording differences based on the Instruction, and provide a final judgment.
\STATE \quad 3. \textbf{Note}:  For mathematics and code problems, please pay attention to the equivalence of the final answer (for example: $\frac{1}{2}$ and 0.5 are equivalent; print("hello") and print('hello') are equivalent in Python).
\end{algorithmic}
\end{algorithm}

\subsubsection{Difficulty Grading}
The corpus lacks explicit difficulty labels, and the base model may not benefit equally from samples of all difficulty levels. To address this, we build a fine-grained difficulty grading system adapted from the strategy used in GLM-5~\cite{zeng2026glm}. The system combines surface-level text features with model-based performance signals: query length and response length serve as basic indicators of sample complexity, while the base model's $\text{pass}@4$ results estimate empirical difficulty. Samples are divided into three levels, i.e., low, medium, and high, based on these signals. Experimental results demonstrate that the optimal $\text{pass}@4$ thresholds are: $[0, 0.25)$ for high difficulty, $[0.25, 0.5)$ for medium difficulty, and $[0.5, 1]$ for low difficulty.

\subsubsection{Classification and Sampling}
To further improve the effectiveness, balance, and scenario diversity of the SFT training data, we conduct task classification and diversity-oriented sampling after distillation, filtering, and difficulty grading.
Due to the diversity of sources, the aggregated data lacks a unified labeling scheme. Therefore, during the classification stage, we adopt a model-assisted workflow. Qwen3-235B-A22B-Instruct~\cite{yang2025qwen3} generates candidate labels, which are combined with human annotations to construct a training set. On this basis, we train a lightweight classifier Qwen3-1.7B to label the task type of all samples. This produces a hierarchical dataset with a clear structure and well-defined categories, which supports more efficient downstream sampling and model training. The specific category classification is shown in Table~\ref{tab:task-taxonomy}. A concrete example of the final data structure is provided in Appendix~\ref{app:data_example}.
During the sampling optimization stage, we apply clustering algorithms to group samples across different task types and difficulty levels, using methods including embedding-based clustering and benchmark-gain-based sampling. While maintaining data quality, this strategy preserves as much scenario diversity and content variation as possible, resulting in a more balanced and representative SFT training corpus.

\begin{table}[t]
\centering
\small
\caption{Classification}
\label{tab:task-taxonomy}
\renewcommand{\arraystretch}{1.3}
\begin{tabularx}{\linewidth}{c l X}
\toprule
\textbf{ID} & \textbf{Main Category} & \textbf{Sub-categories} \\
\midrule
1 & Mathematics                  & Algebra, Geometry, Number Theory, Counting and Probability, Precalculus \\
2 & Code                         & NL-to-PL, PL-to-NL, PL-to-PL, PL-to-Error \\
3 & STEM                         & Physics, Chemistry, Biology, others \\
4 & Instruction Following        & Keyword Restrictions, Length Constraints, Detectable Content, Detectable Format, etc. \\
5 & Natural Language Processing & Text Summarization, Machine Translation, Information Extraction, Rewriting and Polishing, Classification and Tagging, Reading Comprehension \\
6 & Safety \& Alignment          & Safety Refusal, Value Guidance, Self-Awareness \\
7 & Text Creation \& Expression   & Literary Creation, Creative Writing, Role Play \\
\bottomrule
\end{tabularx}
\end{table}

By building an automated corpus processing platform and implementing the end-to-end pipeline described above, we construct a standardized SFT fine-tuning corpus that is high in quality, consistent, diverse, and structurally organized. The resulting corpus contains 1.04M Instruct samples and 2.8M Thinking samples.

\subsection{RL Data Curation}
Following the SFT stage, we construct a 200K-sample verifiable candidate pool for RL training. This candidate pool spans three capability domains where answer correctness can be judged objectively: math (60K samples), code (30K samples), and instruction following (110K samples). The samples are drawn from publicly available RL datasets, including Dolci-Instruct-RL~\cite{olmo2025olmo3}, Eurus-2-RL-Data~\cite{eurus-2-rl-data}, RLVR-IFeval, and IF\_multi\_constraints\_upto5~\cite{pyatkin2025generalizing}. By restricting the candidate pool to verifiable tasks, each sample can receive a reliable reward signal, which is a prerequisite for stable RL training. The actual RL training sets used in later experiments are filtered subsets of this pool: the Instruct RL stage uses 53K samples, while the Reasoning RL stage uses 8K medium-difficulty samples.
The RL data processing pipeline consists of three stages: quality filtering, difficulty grading, and sampling with domain and difficulty balancing. 

In the quality filtering stage, we first perform deduplication across the RL and SFT datasets to remove exact and near-duplicate samples, which prevents the model from over-fitting to repeated patterns and skewing the reward signal. Then, we remove low-quality queries that would introduce noise into RL training. These include garbled text, mixed-language inputs, and incomplete or ill-formed questions; only well-formed, semantically clear queries are retained.

In the difficulty grading stage, we estimate the empirical difficulty of each query by measuring the SFT model's pass rate over five independent responses. For a given query, the SFT model generates five answers, and each answer is checked for correctness using domain-specific verifiers. The verification method is consistent with the SFT-stage procedure. Math answers are validated with the \texttt{math\_verify} tool, code answers are checked through unit tests, and instruction-following responses are evaluated against 54 rule-based verification functions.
The proportion of correct answers across the five trials defines the query's pass rate, which serves as a difficulty proxy: queries with lower pass rates are classified as harder, and those with higher pass rates as easier. 

In the final sampling stage, we select actual training subsets from the filtered and graded candidate pool according to pre-defined domain and difficulty ratios. This balanced sampling strategy ensures that each RL training subset covers diverse capability areas at calibrated difficulty levels, providing the model with well-structured training signals across the full difficulty spectrum.

Before partitioning the corpus into task-specific subsets, we execute an exhaustive global deduplication protocol to guarantee evaluation integrity and prevent benchmark contamination. Specifically, we cross-validate the entire SFT pool and RL candidate pool against all downstream evaluation benchmarks. Any training samples exhibiting high similarity to the test sets are strictly discarded. This rigorous contamination control ensures that the reported performance gains reflect genuine generalization rather than data leakage.

\section{Evaluation Framework}
\label{sec:eval_framework}
\subsection{Evaluation Benchmarks}
To comprehensively and rigorously validate the instruction-following capability and multi-dimensional reasoning performance of our trained models, we conduct automatic quantitative evaluations covering general knowledge, instruction alignment, logical and mathematical reasoning, and code reasoning. All adopted benchmarks are authoritative, challenging, and widely recognized in recent LLM research to ensure fair and credible performance validation. The detailed benchmark categorization corresponding to core model capabilities is presented as follows:
\begin{itemize}
    \item \textbf{General Knowledge and Comprehensive Capability Tasks:} MMLU-Redux~\cite{mmu_redux}, GPQA-Diamond~\cite{gpqa}, C-Eval~\cite{ceval}, and LiveBench (2024-11-25)~\cite{livebench} to evaluate the model’s basic knowledge reserve, cross-domain generalization, and real-world comprehensive problem-solving ability. These benchmarks cover multidisciplinary professional knowledge and open-ended challenging tasks, serving as the fundamental evaluation of the model’s general capability foundation for instruction following and reasoning.

    \item \textbf{Instruction Alignment Tasks:} We employ IFEval~\cite{ifeval} with the strict-prompt accuracy metric as the core evaluation benchmark for instruction-following ability. This benchmark quantitatively assesses whether the model can strictly adhere to complex, fine-grained human instructions, accurately capture key prompt constraints, and generate compliant outputs, which directly reflects the effectiveness of our model’s instruction tuning optimization.

    \item \textbf{Mathematical and Logical Reasoning Tasks:} To fully verify the model’s core reasoning robustness, we select a suite of high-difficulty reasoning benchmarks, including MATH-500~\cite{math500}, AIME'24 and AIME'25 (30 questions for each dataset, with results averaged over 16 sampling runs), ZebraLogic~\cite{zebralogic}, and AutoLogi~\cite{autologi}. These benchmarks cover elementary and advanced mathematical reasoning, symbolic logic reasoning, and deductive reasoning, comprehensively testing the model’s multi-step reasoning, logic derivation, and problem-solving generalization capabilities.

    \item \textbf{Code Reasoning Tasks:} We adopt LiveCodeBench v5 (2024.10--2025.02)~\cite{livecodebench}, a dynamic and challenging code evaluation benchmark, to evaluate the model’s code understanding, logical coding, and algorithm reasoning abilities. All evaluations strictly follow the officially recommended prompt template to guarantee standardized and fair comparison results.
    
\end{itemize}

\subsection{Evaluation Protocol}
For instruct model evaluations in this work, we adopt a unified and fixed generation configuration to eliminate the interference of hyperparameter fluctuations: temperature $=0.7$, top-$p=0.8$, and top-$k=20$, and the maximum output token length is limited to 32,768 tokens to support long responses and complex instruction-following outputs. All instruct models are evaluated under a unified non-thinking protocol.

For reasoning models, we set the sampling temperature to $0.6$, top-$p$ to $0.95$, top-$k$ to $20$, and the maximum output token length to 32,768 tokens. For high-variance mathematical reasoning tasks (AIME'24 and AIME'25), we adopt a 16-sample averaging strategy to stabilize evaluation results and reduce random sampling errors.

\section{Instruct Model Training}
\label{sec:instruct_training}
\subsection{SFT Training for Instruct Models}
SFT dominates the final instruction alignment of modern large language models. Conventional SFT pipelines commonly adopt random sampling or single-criterion data filtering, lacking systematic quantification of data quality and cross-domain mixing principles. Such heuristic strategies easily lead to suboptimal reasoning gain, domain capability conflict, and inevitable alignment degradation, especially severely restricting the upper bound of performance for small-scale (8B) models with limited parameter capacity.

\begin{figure*}[t]
\centering
\includegraphics[width=\textwidth]{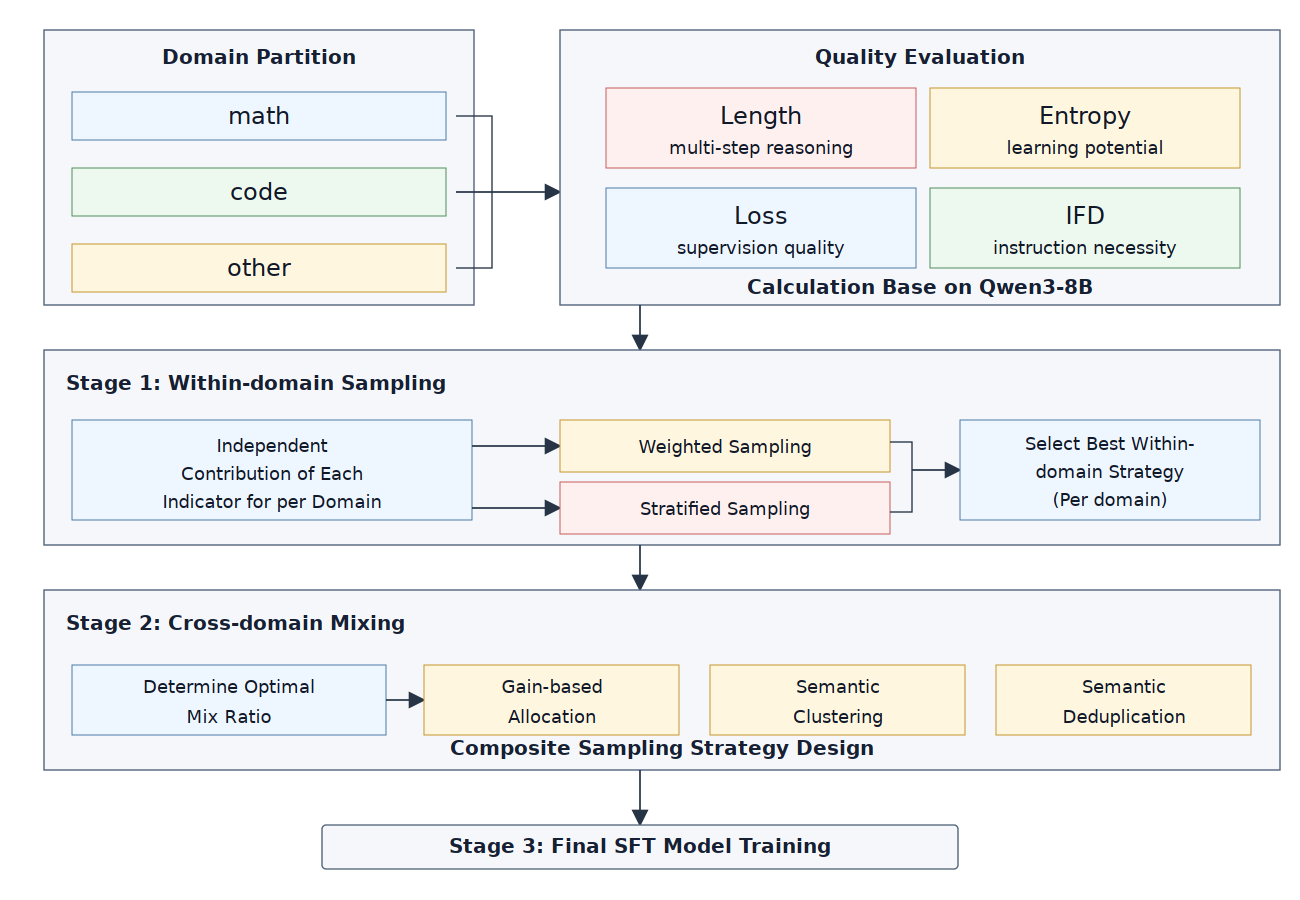}
\caption{Overview of the three-stage SFT data curation pipeline. The pipeline partitions data into three domains, evaluates each example along four quality indicators, determines the optimal within-domain sampling strategy (Stage~1), searches for the best cross-domain mixing ratio (Stage~2), and produces the final SFT model (Stage~3).}
\label{fig:instruct_sft_pipeline}
\end{figure*}

To address these issues, we design a systematic three-stage data curation pipeline for 8B-level instruction tuning. Figure~\ref{fig:instruct_sft_pipeline} provides an overview. From the 1.04M processed Instruct samples, we construct three domain-specific pools for ablation and mixture search---Math (257K samples), Code ($\sim$270K samples), and Other general text (450K samples). We then compute four complementary data quality indicators and progressively determine the optimal within-domain sampling strategy and cross-domain mixing ratio. The remainder of this subsection details each component.

\subsubsection{SFT Technical Route and Details}

\paragraph{Data Quality Indicators}
To comprehensively quantify the training value of instruction-response pairs, we establish a four-dimensional complementary data quality evaluation system covering reasoning integrity, learning potential, data cleanliness, and instruction alignment. All indicators are uniformly calculated with Qwen3-8B-Instruct~\cite{qwen3} as the scoring model to ensure consistent quality calibration; this scorer is separate from the Qwen3-8B-nothink baseline used in our evaluations.

\smallskip
\begin{itemize}
    \item \textbf{Length.} Total token count of the instruction--response pair. Longer sequences typically reflect multi-step reasoning or detailed explanations and are useful for training the model to produce long-form structured responses. Shorter examples, while more efficient to train on, may lack the reasoning depth needed for complex tasks.

    \item \textbf{Entropy.} The predictive entropy $H = -\sum_i p_i \log p_i$ of the response tokens under the target model. High entropy indicates that the model finds the content novel or challenging, pointing to examples with large learning potential. Low entropy signals familiar or templated content that the model has already mastered, offering limited room for improvement.

    \item \textbf{Loss.} Cross-entropy loss of the response tokens under the target model. A low loss means the instruct model considers the response natural and correct; such examples provide clean supervision but may be too easy to drive further learning. A high loss often flags unnatural or erroneous responses and potential label noise.

    \item \textbf{IFD (Instruction-Follow Difficulty).} Following~\cite{li-etal-2024-quantity}, we define $\text{IFD} = \mathcal{L}(r \mid q) \,/\, \mathcal{L}(r)$, the ratio of instruction-conditioned loss to direct (unconditional) response loss. When IFD $<$ 1, the instruction helps the model predict the response---the example exhibits strong instruction--response alignment. When IFD $>$ 1, adding the instruction makes prediction \emph{harder}, indicating a weak or misaligned instruction--response link. Intuitively, low-IFD examples provide clean supervision for format-following, while high-IFD examples force the model to bridge a larger gap between instruction and response, potentially building reasoning capability.
\end{itemize}

Among these, IFD serves as our primary selection criterion. We adopt the original definition from~\cite{li-etal-2024-quantity}: $\text{IFD} = \mathcal{L}(r \mid q) \,/\, \mathcal{L}(r)$. We also experimented with a subtraction-based variant $\mathcal{L}(r) - \mathcal{L}(r \mid q)$, but found it biased toward short responses: with few output tokens, $\mathcal{L}(r)$ is high while $\mathcal{L}(r \mid q)$ drops sharply, producing spuriously favorable scores; as length increases, later tokens become predictable from context, pulling $\mathcal{L}(r)$ down through averaging. The ratio form removes this confound and consistently outperformed subtraction in our experiments (Section~\ref{sec:sft_findings}).

\paragraph{Three-Stage Pipeline}
We propose a principled three-stage SFT pipeline, including within-domain adaptive sampling, cross-domain gain-based mixing, and final model consolidation, to progressively optimize data quality and domain distribution.

\textbf{Stage 1: Within-domain sampling.}
For each domain independently, we fix a training budget (Math: 100K, Code: 100K, Other: 150K examples). We first conduct \emph{single-indicator experiments}: sample the top or bottom $K$ examples by each indicator, train a model, and evaluate on a fixed benchmark suite. This reveals the directional effect of each indicator. Based on these results, we design two composite sampling strategies:

\begin{itemize}
    \item \textbf{Weighted sampling:} For each indicator $m_i$, compute its average gain $g_i$ over target benchmarks relative to the Qwen3-8B-nothink baseline, yielding normalized weights $w_i = g_i / \sum_j g_j$. Each example receives a score $s = \sum_i w_i \cdot r_i$ where $r_i$ is its percentile rank on indicator $i$. The top-$K$ examples by $s$ are selected.

    \item \textbf{Stratified sampling:} For each indicator $m_i$, compute its \emph{contribution score} as the performance drop when removing $m_i$-selected data from the mix. Allocate the total budget $K$ across indicators proportionally to these scores (with smoothing for negative contributions), then select the top-ranked examples within each indicator's allocation without overlap across allocations.
\end{itemize}

\textbf{Stage 2: Cross-domain mixing.}

Fixing the best within-domain strategy from Stage~1, we determine the optimal mixing ratio across domains. We compare three methods: (1)~\emph{Gain-based}: allocate examples proportionally to each domain's average benchmark improvement over baseline; (2)~\emph{Semantic clustering}: embed data, cluster into $K{=}5000$ semantic clusters, allocate proportionally to distinct cluster count per domain; (3)~\emph{Semantic deduplication}: deduplicate directly at the semantic level before mixing. Gain-based mixing (Math 44K / Code 17K / Other 89K, 150K total) achieved the highest average score (60.99, $+$5.98 over baseline) and was selected as the final SFT model. Full evaluation results are reported in Section~\ref{sec:evaluation}.

\textbf{Stage 3: Final model.}
The winning configuration is trained as the final SFT model, with full evaluation results reported in Section~\ref{sec:evaluation}.

\paragraph{Training Configuration}

All models are fine-tuned from Qwen3-8B-base~\cite{qwen3} with fixed hyperparameters: learning rate $5{\times}10^{-5}$, global batch size 1M tokens, 3 epochs, maximum sequence length 32K tokens, using Swift with Zero2 optimization, example packing, and Liger Kernel. The evaluation suite remained fixed throughout all ablation experiments.

\subsubsection{SFT Ablation Studies and Key Findings}
\label{sec:sft_findings}

\paragraph{Single-Indicator Directional Effects}

We sample the top or bottom 100K Math-domain examples by each indicator in isolation. Three patterns emerge: (1)~\textbf{Length} is the strongest single indicator---max-length sampling yields AIME'24 $+9.58$ and AIME'25 $+10.83$ over baseline, while min-length degrades both. (2)~\textbf{IFEval exhibits the inverse preference}---short, low-entropy, low-loss, and low-IFD examples all preserve IFEval better than their counterparts. (3)~\textbf{Division-form IFD outperforms subtraction-form IFD}---max IFD(/) achieves AIME'25 $+12.08$ while max IFD($-$) reaches only $+2.50$, consistent with our theoretical analysis.

\paragraph{Ablation: Independent Contribution of Each Indicator}

To isolate each indicator's marginal contribution, we start from a four-indicator weighted configuration (weights: Length~1.0, Loss~0.81, IFD~0.81, Entropy~0.69, derived from benchmark gains) and remove one indicator at a time (Table~\ref{tab:ablation}).

\begin{table*}[t]
\centering
\caption{Removing each indicator from the Math-domain weighted strategy. Arrows show change vs.\ the full weighted configuration.}
\label{tab:ablation}
\small
\setlength{\tabcolsep}{5pt}
\begin{tabular}{@{}lcccccc@{}}
\toprule
\textbf{Configuration} & \textbf{AIME'24} & \textbf{AIME'25} & \textbf{MATH-500} & \textbf{IFEval} & \textbf{ZebraLogic} & \textbf{LCB v5} \\
\midrule
Qwen3-8B-nothink      & 29.17 & 17.50 & 83.35 & 82.62 & 25.60 & 23.65 \\
\midrule
Weighted (4-indicator) & 42.50 & 29.17 & 90.28 & 76.52 & 26.30 & 20.36 \\
$-$ Length             & 37.50 $\downarrow$5.00 & 26.25 $\downarrow$2.92 & 90.35 & 76.34 & 22.50 $\downarrow$3.80 & 20.66 \\
$-$ Entropy            & 41.67 $\downarrow$0.83 & 29.17~~~0.00 & 90.77 & 76.16 & 26.70 & 21.66 \\
$-$ IFD                & 42.08 $\downarrow$0.42 & 28.75 $\downarrow$0.42 & 90.33 & 74.49 $\downarrow$2.03 & 22.60 $\downarrow$3.70 & 21.56 \\
$-$ Loss               & 38.75 $\downarrow$3.75 & 26.67 $\downarrow$2.50 & 90.85 & 76.16 & 25.60 $\downarrow$0.70 & 22.75 \\
\bottomrule
\end{tabular}
\end{table*}

The ablation establishes \textbf{Length $\gg$ Loss $>$ IFD $\approx$ Entropy} for math reasoning. Removing Length causes the largest drops (AIME'24 $-$5.0, AIME'25 $-$2.92). Removing Loss has the second-largest effect, indicating the instruct model's confidence signal captures quality orthogonal to length. Removing IFD primarily damages IFEval ($-$2.03), suggesting IFD is the indicator most responsible for preserving instruction following within math-heavy data. Entropy has the smallest independent effect, largely subsumed by Length.

\paragraph{Finding 1: Math and code demand opposite data distributions.}

Table~\ref{tab:math_vs_code} contrasts single-indicator results across Math and Code domains.

\begin{table*}[t]
\centering
\caption{Opposite indicator preferences between Math and Code domains. $\Delta$ relative to Qwen3-8B-nothink.}
\label{tab:math_vs_code}
\small
\setlength{\tabcolsep}{9pt}
\begin{tabular}{@{}lcccc@{}}
\toprule
& \multicolumn{2}{c}{\textbf{Math domain (100K)}} & \multicolumn{2}{c}{\textbf{Code domain (100K)}} \\
\cmidrule(lr){2-3} \cmidrule(lr){4-5}
\textbf{Sampling} & \textbf{AIME'24 $\Delta$} & \textbf{IFEval $\Delta$} & \textbf{LCB v5 $\Delta$} & \textbf{AIME'24 $\Delta$} \\
\midrule
Max Length     & \textbf{$+$9.58} & $-$5.36 & $+$0.30 & $+$3.33 \\
Max Entropy    & $+$5.83 & $-$4.06 & $-$0.90 & $+$1.66 \\
Max IFD(/)     & $+$3.75 & $-$2.03 & $-$5.98 & $-$2.92 \\
\midrule
Min Entropy    & $+$2.50 & $-$2.40 & \textbf{$+$1.50} & $-$7.09 \\
Min IFD(/)     & $+$1.66 & $-$0.92 & $+$0.90 & $-$5.00 \\
\bottomrule
\end{tabular}
\end{table*}

The pattern is symmetric: \textbf{Math reasoning benefits from high entropy and high IFD; code generation benefits from low entropy and low IFD.} Under our IFD definition, high IFD means the instruction provides weaker guidance---forcing the model to independently reconstruct the reasoning path, which plausibly strengthens math capability. Low IFD means the instruction is tightly coupled to the response, providing clean supervision for format-sensitive tasks like code generation. Optimizing exclusively for one domain damages the other: Math-weighted sampling achieves AIME'24 42.50 but LiveCodeBench drops to 20.36; the Code-optimal configuration (LiveCodeBench 25.15) sees AIME'24 fall to 28.33. We attribute this zero-sum conflict to 8B-scale capacity constraints---deterministic generation and flexible reasoning impose opposing pressure on the same parameter budget; whether this generalizes to larger models remains open.

\paragraph{Finding 2: Long-form general text transfers to reasoning.}

The Other domain ($\sim$450K examples, no dedicated math or code) yields a surprising result: sampling 150K examples purely by maximum length enables the model to surpass Qwen3-8B-nothink on AIME'24 (37.50 vs.\ 29.17), AIME'25 (25.42 vs.\ 17.50), MATH-500 (89.00 vs.\ 83.35), and ZebraLogic (64.10 vs.\ 25.60, $+$38.5)---all without a single math training example. The cost is an alignment tax (IFEval $-$6.46, LiveCodeBench $-$1.49). We hypothesize that long-form, logically structured text teaches sustained reasoning chains that transfer to formal tasks. We caution that some Other-domain examples may resemble ZebraLogic puzzles; quantifying leakage versus genuine transfer is left to future work.

\paragraph{Finding 3: Cross-domain mixing dynamics.}

Fixing the best within-domain strategies, we evaluated mixing methods at 150K total examples. Three conclusions emerge: (1)~Gain-based mixing (Math 44K / Code 17K / Other 89K) achieves the highest average score (60.99, $+$5.98). (2)~Semantic deduplication underperforms in our setting (LiveCodeBench 19.46, $-$4.19), suggesting that moderate near-duplicate structure may provide useful training signals. (3)~The alignment tax is recoverable: adding dedicated IF examples restores IFEval to 82.26. Subsequent refinements with stratified$+$weighted sampling further improve code and instruction-following at a modest average cost. Increasing the code proportion raised MATH-500 from 88.60 to 91.00 without improving LiveCodeBench, indicating code data transfers structured reasoning patterns to math.

\paragraph{The underrated factor: correctness filtering.}

A simple correctness cross-validation filter---removing examples whose responses fail answer verification---produced gains rivaling sophisticated sampling strategies (Table~\ref{tab:correctness}). The unfiltered configuration (IFD $+$ Length weighted, 100K) contained 7,106 examples (7.1\%) that failed cross-validation; removing them improved AIME'25 by $+$7.09, AIME'24 by $+$3.75, and IFEval by $+$2.21, surpassing Qwen3-8B-nothink on IFEval without dedicated IF data. This suggests \textbf{data correctness is the first-order filter; sampling strategy is a second-order optimization} on top of already-verified data.

\begin{table}[t]
\centering
\caption{Correctness cross-validation filtering: removing 7.1\% failed examples.}
\label{tab:correctness}
\small
\begin{tabular}{@{}lccc@{}}
\toprule
\textbf{Benchmark} & \textbf{Before filter} & \textbf{After filter} & \textbf{$\Delta$} \\
\midrule
IFEval          & 81.15 & \textbf{83.36} & $+$2.21 \\
AIME'24          & 28.75 & 32.50 & $+$3.75 \\
AIME'25          & 22.08 & 29.17 & $+$7.09 \\
LiveCodeBench v5 & 25.75 & 26.95 & $+$1.20 \\
MATH-500        & 86.67 & 86.05 & $-$0.62 \\
\bottomrule
\end{tabular}
\end{table}

\subsubsection{SFT Model Results}
\label{sec:evaluation}

Table~\ref{tab:main_results} presents the main results.

\begin{table*}[t]
\centering
\caption{Non-thinking evaluation results. Bold indicates the best score in each row. AIME scores are 16-sample averages throughout.}
\label{tab:main_results}
\small
\setlength{\tabcolsep}{5pt}
\begin{tabular}{@{}lccccc@{}}
\toprule
\textbf{Benchmark} & \textbf{Qwen3-8B-nothink} & \textbf{NebulaExp-Ins-SFT} & \textbf{Math-only} & \textbf{Code-only} & \textbf{Other-only} \\
\midrule
\multicolumn{6}{c}{\textit{General Tasks}} \\
\midrule
MMLU-Redux      & 86.23 & 84.07 & 84.67 & \textbf{87.27} & 83.50 \\
GPQA-Diamond    & 47.98 & 52.53 & 52.53 & 47.98 & \textbf{55.05} \\
C-Eval          & 79.05 & \textbf{82.91} & 81.65 & 81.87 & 82.76 \\
LiveBench       & \textbf{53.10} & 46.50 & 43.70 & 44.20 & 43.90 \\
\midrule
\multicolumn{6}{c}{\textit{Alignment}} \\
\midrule
IFEval          & \textbf{82.62} & 77.08 & 76.52 & 79.11 & 75.79 \\
\midrule
\multicolumn{6}{c}{\textit{Math \& Text Reasoning}} \\
\midrule
MATH-500        & 83.35 & 88.60 & 90.28 & 87.00 & \textbf{91.00} \\
AIME'24         & 29.17 & \textbf{44.17} & 42.50 & 33.33 & 31.67 \\
AIME'25         & 17.50 & 28.75 & \textbf{29.17} & 25.83 & 24.58 \\
AutoLogi        & 76.91 & \textbf{81.08} & 75.35 & 70.78 & 77.53 \\
ZebraLogic      & 25.60 & 61.50 & 26.30 & 22.80 & \textbf{63.90} \\
\midrule
\multicolumn{6}{c}{\textit{Coding}} \\
\midrule
LiveCodeBench v5 & 23.65 & 23.65 & 20.36 & 23.35 & 22.75 \\
\midrule
Average         & 55.01 & \textbf{60.99} & 56.64 & 54.87 & 59.31 \\
\bottomrule
\end{tabular}
\end{table*}

\paragraph{General and Knowledge Tasks}

NebulaExp-Ins-SFT improves over Qwen3-8B-nothink on GPQA-Diamond (52.53, $+$4.55) and C-Eval (82.91, $+$3.86), while MMLU-Redux (84.07, $-$2.16) and LiveBench (46.50, $-$6.60) show modest regression. The domain-only models reveal the trade-off: Other-only achieves the highest GPQA (55.05) through broad factual coverage but suffers on MMLU-Redux (83.50), while Code-only best preserves MMLU-Redux (87.27) at the cost of weaker reasoning. Our cross-domain mix strikes a balance, maintaining competitive knowledge scores while delivering large reasoning gains.

\smallskip

\noindent\textbf{LiveBench Subcategory Analysis.}
LiveBench spans six subcategories. Table~\ref{tab:livebench_sub} presents the subcategory breakdown from our within-domain Math experiments, illustrating the capability trade-offs induced by different sampling strategies.

\begin{table*}[t]
\centering
\caption{LiveBench subcategory scores under different Math-domain sampling strategies (non-thinking). All use 100K Math-domain examples.}
\label{tab:livebench_sub}
\small
\setlength{\tabcolsep}{7pt}
\begin{tabular}{@{}lccccccc@{}}
\toprule
\textbf{Subcategory} & \textbf{Qwen3-8B-nothink} & \textbf{Length-mix} & \textbf{maxL} & \textbf{maxE} & \textbf{minLoss} & \textbf{maxIFD/} & \textbf{Weighted} \\
\midrule
Math            & 54.8 & 60.0 & \textbf{60.3} & 58.7 & 56.9 & 59.2 & 58.7 \\
Reasoning       & 51.8 & \textbf{63.2} & 56.7 & 50.5 & 51.2 & 55.8 & 49.0 \\
Coding          & 43.9 & 35.1 & 35.2 & \textbf{40.8} & 38.1 & 37.4 & 37.8 \\
Data Analysis   & 49.0 & 38.3 & 37.0 & 34.7 & \textbf{44.4} & 36.4 & 38.8 \\
Inst.\ Following & 73.5 & 58.6 & 60.2 & \textbf{64.7} & 63.9 & 60.5 & 60.7 \\
Language        & 45.9 & 11.1 & 13.1 & 12.1 & 13.5 & \textbf{17.7} & 17.2 \\
\midrule
Total           & \textbf{53.1} & 44.4 & 43.8 & 43.6 & 44.7 & 44.5 & 43.7 \\
\bottomrule
\end{tabular}
\end{table*}

The decomposition reveals a consistent trade-off: strategies that maximize Math and Reasoning (long Length, high Entropy) systematically degrade Coding, Instruction Following, and Language. Length-mix (80\% longest $+$ 20\% random) achieves the highest Reasoning (63.2, $+$11.4) but Language collapses to 11.1 ($-$34.8). Our final cross-domain mix resolves this tension: the Other domain supplies long-form language data that restores Language and Instruction Following, while Math and Code contribute targeted reasoning improvements. The net effect is a LiveBench total (46.50) that avoids the extreme subcategory collapse seen in single-domain sampling, though it remains below the Qwen3-8B-nothink baseline (53.10).

\paragraph{Alignment and Instruction Following}

IFEval (77.08) trails Qwen3-8B-nothink (82.62) by $-$5.54, reflecting the ``alignment tax'' inherent in our data mix: 59.4\% of training examples come from the Other domain, selected for reasoning depth rather than instruction-following precision. All single-domain models score lower on IFEval, confirming that dedicated math, code, or general-text data alone is insufficient for instruction following. Our ablation experiments (Section~\ref{sec:sft_findings}) demonstrate this tax is recoverable: adding dedicated IF examples restores IFEval to 82.26. We report the gain-based configuration here as it maximizes aggregate performance; IF-augmented variants are available for scenarios prioritizing instruction following.

\paragraph{Math and Text Reasoning}

NebulaExp-Ins-SFT demonstrates substantial gains across math and reasoning: AIME'24 44.17 ($+$15.00), AIME'25 28.75 ($+$11.25), MATH-500 88.60 ($+$5.25), and AutoLogi 81.08 ($+$4.17). ZebraLogic shows the most dramatic improvement (61.50, $+$35.90 over Qwen3-8B-nothink), driven by long-form reasoning patterns transferred from the Other domain (Finding~2, Section~\ref{sec:sft_findings}).

The single-domain models reveal the ceiling of single-objective optimization: Math-only reaches AIME'24 42.50 but pays a heavy price in IFEval (76.52) and LiveCodeBench (20.36). Other-only achieves ZebraLogic 63.90 and MATH-500 91.00 purely through long-form text, without dedicated math data. Our cross-domain configuration trades some of this peak performance for balanced gains, achieving the highest average score (60.99) while maintaining competitive scores across all categories. The ablation studies (Section~\ref{sec:sft_findings}) show math capability is most sensitive to Length and Entropy---dimensions where the Other domain (59.4\% of the training mix) provides a complementary signal through long-form text.

\paragraph{Coding}

LiveCodeBench v5 remains at parity with Qwen3-8B-nothink (23.65). Code data constitutes only 11.2\% of the training mix (16,806 examples), yet the model maintains baseline coding performance through our stratified sampling strategy (low-Entropy $+$ low-IFD), which our ablation studies identified as optimal for code generation. The Code-only single-domain model achieves 23.35, indicating the cross-domain mix largely preserves code capability. Our later configurations with additional IF data and stratified sampling show that increasing the code proportion and adding IF data can push LiveCodeBench substantially higher (up to 26.95), at a modest cost to aggregate average.

\paragraph{Summary}

NebulaExp-Ins-SFT achieves an average score of 60.99 ($+$5.98 over Qwen3-8B-nothink). Gains are broad-based: AIME'24 $+$15.00, AIME'25 $+$11.25, ZebraLogic $+$35.90, AutoLogi $+$4.17, GPQA $+$4.55, and C-Eval $+$3.86. The main regressions are IFEval ($-$5.54, the ``alignment tax'') and LiveBench ($-$6.60), both attributable to the reasoning vs.\ instruction-following trade-off identified in our ablation studies. The alignment tax is recoverable through targeted IF data augmentation (IFEval 82.26 with dedicated IF data). Trained via a systematic three-stage data curation pipeline on the Dolci corpus, NebulaExp-Ins-SFT outperforms Qwen3-8B-nothink while providing a transparent, reproducible recipe for SFT data selection.

\subsection{Reinforcement Learning for Instruct Models}

Following SFT, we conduct large-scale mixed-domain RL training using GRPO~\cite{shao2024deepseekmath} on a 53K-sample filtered training subset selected from the 200K RL candidate pool. This actual training subset spans three domains: math (27.5K samples), instruction following (14.5K samples), and code (11K samples). All design decisions in this section---including data selection criteria, training hyperparameters, and epoch budget---are informed by controlled ablation studies reported in Section~\ref{sec:rl_ablation}. We organize this section as follows: first we describe the RL training subset and methodology, then present the ablation experiments that guided our design choices, and finally report the final model trained under the optimal configuration and its benchmark performance.

\subsubsection{RL Technical Route and Details}
We describe the RL data selection and training configuration.

\textbf{Data selection.}
Math samples are drawn from Dolci-Instruct-RL, code samples are sourced from Eurus-2-RL-Data-code, and instruction-following samples come from RLVR-IFeval. All of them are selected with the constraint $\text{pass\_rate} \notin \{0, 1\}$. The pass\_rate filter is motivated by our difficulty ablation (Section~\ref{sec:rl_ablation}), which shows that samples with a pass rate of 0 (trivially unsolvable) or 1 (already mastered) contribute negligible advantage signals and removing them yields a measurable gain in average score (60.24 vs.\ 59.89). After filtering, the domain composition of the 53K training subset reflects the availability of high-quality verifiable data in each category, with math and instruction-following samples together accounting for 79\% of the subset.

\textbf{Training configuration.}
The base model for this stage is NebulaExp-Ins-SFT (average score 60.99). Training uses a rollout size of 5, a maximum response length of 16,384 tokens, and a sequence-parallel size of 8. The model is trained for up to 6 epochs with KL loss enabled. These settings follow the findings of our training-technique ablation: retaining KL loss stabilizes convergence without degrading final performance (60.24 vs.\ 60.07 without KL), relaxing the strict on-policy constraint accelerates training, and a larger rollout size yields stronger results at increased computational cost. We select rollout size 5 as a practical trade-off between performance and training time; the ablation experiments with rollout size 10 achieve higher scores (61.11) but increase total training time by approximately 70\%.

\subsubsection{RL Ablation Studies and Key Findings}
\label{sec:rl_ablation}

We conduct two sets of controlled ablation experiments to assess the impact of sample difficulty and training techniques on RL performance. To control training time for individual experiment, each ablation experiment uses a reduced corpus of 10K samples (spanning math, code, and instruction following at a ratio of 47:30:23) and trains for 4 epochs.

\textbf{Effect of sample difficulty.}
The difficulty of each sample is determined by the base model's pass rate over multiple rollouts. For math and instruction-following samples, we use five rollouts; for code, we use two rollouts. Correctness is judged by domain-specific verifiers (math\_verify for math, unit tests for code, and 25 rule-based checks for instruction following). Samples with a pass rate of 0 are trivially unsolvable by the base model, and those with a pass rate of 1 are already mastered; both categories provide negligible advantage signals for RL.
Table~\ref{tab:rl_difficulty_ablation} compares five configurations: training on all samples (avg.\ 59.89); removing samples with pass rates of 0 or 1 (avg.\ 60.24); retaining only easy samples with pass rate $\in [0.5, 1)$ (avg.\ 59.44); retaining only hard samples with pass rate $\in (0, 0.5)$ (avg.\ 60.74); and curriculum learning (CL), where the model is first trained on easy samples and then fine-tuned on hard samples (avg.\ 60.41). The results show a clear ordering: hard-only data yields the best performance, followed by curriculum learning, mixed data with 0/1 removal, unfiltered data, and easy-only data. Removing samples at the extremes (pass rates of 0 and 1) alone brings a measurable gain, confirming that these samples contribute little to RL training.
The finding that hard-only training outperforms curriculum learning differs from the prevailing observation in the literature, where curriculum learning typically achieves better results. We attribute this discrepancy to the relatively small model scale (8B parameters), at which the model may benefit more from concentrated exposure to challenging samples than from progressive difficulty scheduling.

\begin{table*}[t]
\centering
\caption{RL evaluation results for sample difficulty ablations.}
\label{tab:rl_difficulty_ablation}
\small
\setlength{\tabcolsep}{5pt}
\begin{tabular}{@{}lcccccc@{}}
\toprule
\textbf{Benchmark} & \textbf{Qwen3-8B-nothink} & \textbf{All samples} & \textbf{Pass rate $\in (0, 1)$} & \textbf{Pass rate $\in [0.5, 1)$} & \textbf{Pass rate $\in (0, 0.5)$} & \textbf{CL}\\
\midrule
\multicolumn{7}{c}{\textit{General Tasks}} \\
\midrule
MMLU-Redux      & \textbf{86.23} & 84.60 & 84.40  & 84.50  & 84.37 & 84.43 \\
GPQA-Diamond    & 47.98 & 52.02 & \textbf{53.54}  & 49.49 & 52.02 & 51.01 \\
C-Eval          & 79.05 & 82.32 & \textbf{83.66}  & 82.17 & 81.95 & 83.21 \\
LiveBench       & \textbf{53.10} & 45.60 & 45.30  & 45.30 & 46.60 & 46.50 \\
\midrule
\multicolumn{7}{c}{\textit{Alignment}} \\
\midrule
IFEval          & 82.62 & 76.89 & 79.85  & \textbf{80.22} & 78.37 & 78.56 \\
\midrule
\multicolumn{7}{c}{\textit{Math \& Text Reasoning}} \\
\midrule
MATH-500        & 83.35 & 89.40 & 91.00  & \textbf{91.80} & 90.20 & 91.60 \\
AIME'24         & 29.17 & 40.83 & 41.88  & 40.21 & 42.50 & \textbf{46.67} \\
AIME'25         & 17.50 & 31.87 & 26.84  & 29.33 & \textbf{33.75} & 28.33 \\
AutoLogi        & 76.91 & 79.04 & 80.49  & 80.04 & \textbf{80.88} & 78.82 \\
ZebraLogic      & 25.60 & \textbf{52.60} & 51.70  & 51.60 & 52.00 & 51.10 \\
\midrule
\multicolumn{7}{c}{\textit{Coding}} \\
\midrule
LiveCodeBench v5 & 23.65 & 23.65 & 23.95  & 19.16 & \textbf{25.45} & 24.25 \\
\midrule
Average         & 55.01 & 59.89 & 60.24  & 59.44 & \textbf{60.74} & 60.41 \\
\bottomrule
\end{tabular}
\end{table*}

\begin{figure*}[t]
\centering
\includegraphics[width=\textwidth]{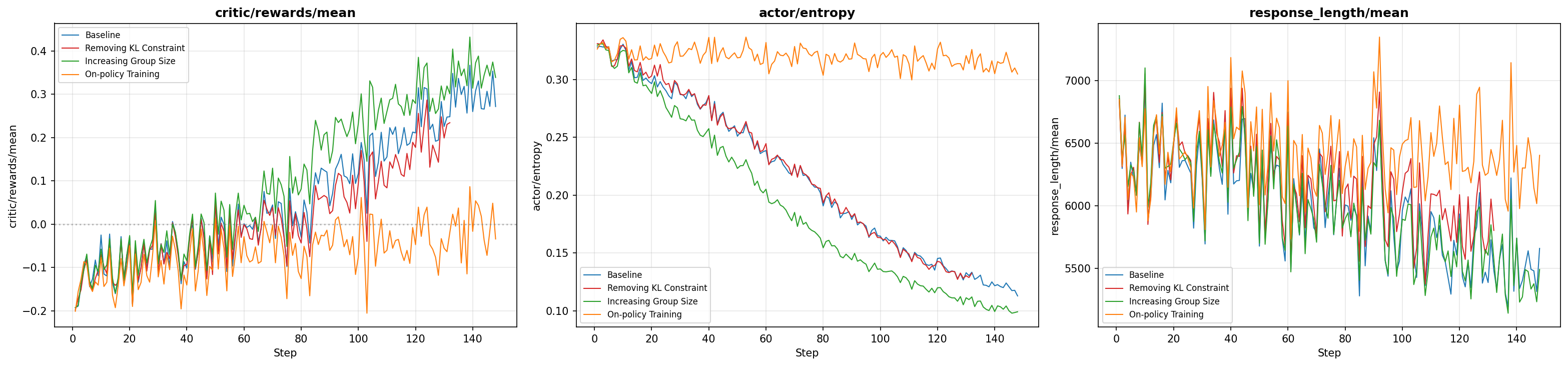}
\caption{Training dynamics of the four training technique ablation experiments. The three panels show (a) critic/rewards/mean, (b) actor/entropy, and (c) response\_length/mean as a function of training steps. The Baseline configuration uses GRPO with rollout size 5 and KL loss. Removing KL Constraint has minimal impact on reward convergence, consistent with its negligible effect on final benchmark scores. Increasing Group Size (rollout size 10) yields higher rewards and faster entropy decay, indicating more efficient per-step learning, though at the cost of 70\% longer training time. On-policy Training exhibits notably slower convergence in reward growth, entropy reduction, and response length increase, corroborating its lowest final average score.}
\label{fig:rl_ablation_curves}
\end{figure*}

\textbf{Effect of training techniques.}
We evaluate three training modifications against the GRPO baseline (rollout size 5, KL loss enabled, 4 epochs, train batch size 256, PPO mini-batch size 64), which achieves an average score of 60.24. The variants are: (a) disabling the KL loss and removing KL from the reward (avg.\ 60.07); (b) strict on-policy training with the PPO mini-batch size set equal to the train batch size of 256, so that importance sampling ratios are always 1 (avg.\ 59.65); and (c) increasing the rollout size from 5 to 10 (avg.\ 61.11). Table~\ref{tab:rl_trick_ablation} summarizes the results.
The conclusions are as follows: 1) Removing the KL constraint has a negligible effect on final performance and only slightly slows the rate of model convergence, indicating that the KL penalty is not a critical regularizer under this setup. 2) Strict on-policy training prevents the amplification of gradients for high-reward tokens and leads to slower convergence in terms of model updates, entropy reduction, and reward growth, resulting in the lowest final score. 3) Increasing the group size to 10 yields the strongest result, as larger groups reduce the probability of all-zero advantage batches and improve data utilization. However, rollout generation accounts for roughly 70\% of the total training time, so this gain comes with a substantial increase in computational cost. Figure~\ref{fig:rl_ablation_curves} visualizes these training dynamics across all four variants.
Based on these findings, the configuration we adopt for large-scale training is: filtering out samples with pass rates of 0 or 1, retaining the KL loss for stability and compatibility with later large-scale training despite its small final-score impact, relaxing the strict on-policy constraint, and using a moderate rollout size of 5 as a practical trade-off between performance and compute.

\begin{table*}[t]
\centering
\caption{RL evaluation results for training techniques ablations.}
\label{tab:rl_trick_ablation}
\small
\setlength{\tabcolsep}{5pt}
\begin{tabular}{@{}lccccc@{}}
\toprule
\textbf{Benchmark} & \textbf{Qwen3-8B-nothink} & \textbf{Baseline} & \textbf{Removing KL Constraint} & \textbf{On-policy Training} & \textbf{Increasing Group Size} \\
\midrule
\multicolumn{6}{c}{\textit{General Tasks}} \\
\midrule
MMLU-Redux      & \textbf{86.23} & 84.40 & 84.73  & 84.93  & 83.83 \\
GPQA-Diamond    & 47.98 & 53.54 & 50.00  & 53.54  & \textbf{55.56} \\
C-Eval          & 79.05 & \textbf{83.66} & 82.76  & 83.28  & 83.51 \\
LiveBench       & \textbf{53.10} & 45.30 & \textbf{46.80}  & 46.30  & 46.70 \\
\midrule
\multicolumn{6}{c}{\textit{Alignment}} \\
\midrule
IFEval          & 82.62 & \textbf{79.85} & 79.67  & 78.93  & 79.67 \\
\midrule
\multicolumn{6}{c}{\textit{Math \& Text Reasoning}} \\
\midrule
MATH-500        & 83.35 & 91.00 & \textbf{91.80}  & 89.80  & 91.60 \\
AIME'24         & 29.17 & 41.88 & 42.08  & \textbf{44.38}  & 39.79 \\
AIME'25         & 17.50 & 26.84 & 25.83  & 23.12  & \textbf{32.08} \\
AutoLogi        & 76.91 & 80.49 & 79.49  & \textbf{79.97}  & 79.91 \\
ZebraLogic      & 25.60 & 51.70 & 53.10  & 48.80  & \textbf{55.30} \\
\midrule
\multicolumn{6}{c}{\textit{Coding}} \\
\midrule
LiveCodeBench v5 & 23.65 & 23.95 & \textbf{24.55}  & 23.05 & 24.25 \\
\midrule
Average         & 55.01 & 60.24 & 60.07  & 59.65  & \textbf{61.11} \\
\midrule
Training time   & -     & T     & T     & T       & 1.7T \\
\bottomrule
\end{tabular}
\end{table*}

\subsubsection{RL Model Results}

\begin{figure*}[t]
\centering
\includegraphics[width=\textwidth]{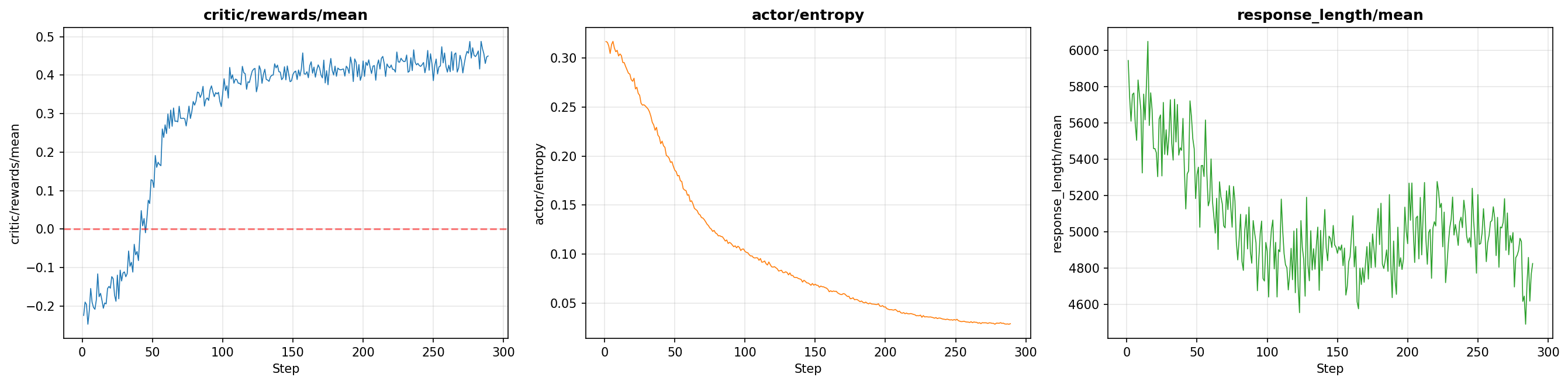}
\caption{Training curves of the final Instruct RL model on the 53K training subset. The three panels show (a) critic/rewards/mean, (b) actor/entropy, and (c) response\_length/mean over the full training run. }
\label{fig:rl_training_curves}
\end{figure*}

Guided by the ablation findings above, we train the final RL model on the full 53K training subset under the optimal configuration: samples with pass rates of 0 or 1 are removed, KL loss is retained, the on-policy constraint is relaxed (PPO mini-batch size 64 vs.\ train batch size 256), and a rollout size of 5 is used as a practical trade-off between performance and training cost. The base model is NebulaExp-Ins-SFT (average score 60.99), and training proceeds for up to 6 epochs with early stopping applied when the reward converges (at approximately 300 steps). Figure~\ref{fig:rl_training_curves} shows the training dynamics of the final model.

Table~\ref{tab:rl_large_scale} reports the full benchmark results. After RL training, the average score improves from 60.99 to 61.85 ($+$0.86). We analyze the results by capability domain below.

\begin{itemize}
    \item \noindent\textit{General tasks.}
RL training yields consistent improvements across knowledge-oriented benchmarks. MMLU-Redux rises from 84.07 to 85.50 ($+$1.43), recovering part of the gap with the Qwen3-8B-nothink baseline (86.23). GPQA-Diamond improves from 52.53 to 53.54 ($+$1.01), establishing a new best across all configurations. C-Eval increases from 82.91 to 83.73 ($+$0.82). LiveBench shows a modest gain from 46.50 to 47.00, indicating that RL on verifiable tasks provides limited benefit for the diverse, open-ended evaluation scenarios captured by LiveBench.
    \item \noindent\textit{Alignment.}
IFEval exhibits the largest single gain among all benchmarks, rising from 77.08 to 80.96 ($+$3.88). We attribute this gain to the inclusion of 14.5K verifiable instruction-following samples in the RL corpus, which provide targeted reward signals for constraint satisfaction.
    \item \noindent\textit{Math and reasoning.}
MATH-500 improves from 88.60 to 91.20 ($+$2.60), surpassing the previous best (91.00 from the Other-only SFT variant) and closing in on the saturated regime. AIME'24 reaches 45.62 ($+$1.45 over SFT), the highest score across all configurations. AIME'25 shows a marginal decline (28.75 to 28.33, $-$0.42), indicating that the RL reward signal may be less effective for the most challenging math problems where the model struggles to produce any correct trajectory within the rollout budget. AutoLogi declines from 81.08 to 78.14 ($-$2.94)---a domain where RL on math-heavy data does not naturally transfer. ZebraLogic remains stable (61.50 to 61.80, $+$0.30), suggesting the long-form reasoning capabilities acquired during SFT are largely preserved.
    \item \noindent\textit{Coding.}
LiveCodeBench v5 increases from 23.65 to 24.55 ($+$0.90), a modest improvement consistent with the relatively small proportion of code data (11K samples, 20.8\%) in the RL corpus. Unlike math and instruction following, where answer verification is relatively mature, code generation rewards rely on unit-test pass rates which provide sparser and noisier signals. The ablation experiments (Section~\ref{sec:rl_ablation}) confirm that code improvements are attainable with larger code-data allocations and optimized difficulty filtering.
\end{itemize}

\textbf{Summary.}
RL training on verifiable samples delivers broad improvements over the SFT checkpoint, with the largest gains in instruction following (IFEval $+$3.88) and math (MATH-500 $+$2.60, AIME'24 $+$1.45). The improvements are most pronounced in domains well-represented in the RL corpus, while out-of-domain capabilities (AutoLogi, LiveBench) see modest gains or slight regressions. The overall average rises from 60.99 to 61.85, confirming that verifiable RL provides a reliable, though incremental, path to post-SFT improvement at the 8B scale.

\begin{table}[t]
\centering
\caption{RL evaluation results.}
\label{tab:rl_large_scale}
\small
\begin{tabular}{@{}lccc@{}}
\toprule
\textbf{Benchmark} & \textbf{Qwen3-8B-nothink} & \textbf{NebulaExp-Ins-SFT} & \textbf{NebulaExp-Ins-RL} \\
\midrule
\multicolumn{4}{c}{\textit{General Tasks}} \\
\midrule
MMLU-Redux      & \textbf{86.23} & 84.07 & 85.50 \\
GPQA-Diamond    & 47.98 & 52.53 & \textbf{53.54}  \\
C-Eval          & 79.05 & 82.91 & \textbf{83.73}  \\
LiveBench       & \textbf{53.10} & 46.50 & 47.00  \\
\midrule
\multicolumn{4}{c}{\textit{Alignment}} \\
\midrule
IFEval          & \textbf{82.62} & 77.08 & 80.96  \\
\midrule
\multicolumn{4}{c}{\textit{Math \& Text Reasoning}} \\
\midrule
MATH-500        & 83.35 & 88.60 & \textbf{91.20}  \\
AIME'24         & 29.17 & 44.17 & \textbf{45.62}  \\
AIME'25         & 17.50 & \textbf{28.75} & 28.33  \\
AutoLogi        & 76.91 & \textbf{81.08} & 78.14  \\
ZebraLogic      & 25.60 & 61.50 & \textbf{61.80}  \\
\midrule
\multicolumn{4}{c}{\textit{Coding}} \\
\midrule
LiveCodeBench v5 & 23.65 & 23.65 & \textbf{24.55}  \\
\midrule
Average         & 55.01 & 60.99 & \textbf{61.85}  \\
\bottomrule
\end{tabular}
\end{table}

\section{Reasoning Model Training}\label{sec:reasoning_training}

This section presents the post-training pipeline for the Reasoning model, covering SFT, RL, and OPD stages. We investigate how data difficulty, reasoning-chain length, mixture ratios, and training paradigms jointly affect reasoning and general capability, using the same benchmark suite and evaluation protocol described in Section~\ref{sec:eval_framework}.

\subsection{SFT Training for Reasoning Models}

\subsubsection{SFT Technical Route and Details}
The SFT stage is designed as a progressive data-optimization process. We first decouple the major factors that affect SFT performance, including mathematical sampling strategy, multi-domain mixture ratio, and the relationship between data scale, and training epochs. Based on these factors, the final SFT recipe is determined through three sequential steps.

First, math is highly sensitive to sample difficulty, reasoning-chain length, and curriculum design. Therefore, we conduct controlled experiments on math reasoning data to identify an effective sampling strategy. We compare difficulty-based sampling, length-based sampling, data-scale variants, and short-to-long curriculum learning. These experiments show that long reasoning-chain samples and curriculum-style training are more effective than simply increasing the amount of mixed mathematical data. Therefore, the final SFT corpus is constructed by prioritizing high-quality long-chain reasoning samples and using staged training to improve the stability of reasoning-format learning.

Second, after fixing the sampling strategy, we search for a better mixture ratio among different data types. Following the referenced mixture-ratio optimization method, we treat the selection of data proportions as a measurable prediction problem rather than relying only on manual heuristics. Specifically, candidate mixtures are built from math, code, and science data. Proxy training is then used to obtain evaluation signals for each candidate mixture, and a regression model is fitted to estimate the relationship between mixture ratios and evaluation performance. The predicted high-performing mixture is used as the basis for the full SFT data composition, so that the final corpus can balance the proportion of different corpus categories.

Third, after determining the mixture ratio, we further optimize the training data according to the principle of training smaller but higher-quality datasets for more epochs. Inspired by prior work on data filtering and repeated training over compact high-quality corpora, we reduce redundancy and noise through category-level cleaning, deduplication, and filtering, while keeping the optimized mixture ratio as stable as possible. The filtered corpus is then trained for more epochs, which improves training efficiency and reduces the negative impact of noisy or duplicated samples, while allowing the model to repeatedly learn from higher-value reasoning and instruction examples.

Overall, the SFT technical route follows the logic of first optimizing the mathematical sampling strategy, then using this strategy to support mixture-ratio search across data types, and finally applying filtering and multi-epoch training to obtain a stronger SFT model. This design avoids coupling all decisions into a single full-scale training run and provides clearer experimental evidence for each stage of the final SFT recipe.

\subsubsection{SFT Ablation Studies and Key Findings}
We conduct a series of experiments around key factors to validate core assumptions in the data construction process and to provide reusable data processing strategies for subsequent SFT and RL training. This subsubsection is organized according to the logic of data sources and composition, data processing methods, experimental design, and results and conclusions.

\paragraph{Ablation Study on Math Data Difficulty and Sampling Strategy}

\textbf{Data sources and composition:} Mathematical reasoning ability is sensitive to sample difficulty, the length of reasoning chains, and data scale. To analyze the effects of different data factors on Reasoning model training, we construct multiple subsets from 150 thousand mathematical reasoning samples. These include hard, mid, and easy subsets divided by difficulty; long and short subsets divided by reasoning text length; and subsets of different scales, including 50K, 75K, and 150K samples.

\textbf{Data processing method:} This experiment constructs datasets around three sampling strategies. The first is difficulty-based sampling, which divides data into hard, mid, and easy subsets according to problem complexity, in order to evaluate the impact of sample difficulty on complex reasoning ability. The second is length-based sampling, which divides data into long and short subsets according to the length of the answer reasoning process, in order to verify whether long chain-of-thought samples are more beneficial for learning step-by-step reasoning. The third is scale-based sampling, which constructs datasets of different sizes under the same sampling standard to determine whether increasing data volume continues to improve performance. In addition, we design a two-stage curriculum learning experiment from short to long samples to compare single-stage mixed training with staged training.

\textbf{Experimental design:} All experiments use Qwen3-8B-base as the base model and fix the basic training configuration: three epochs, a learning rate of 5e-5, cosine learning-rate annealing, and a context length of 32K. By controlling one variable at a time, the experiments compare the impact of difficulty distribution, sample length, data scale, and curriculum learning strategy on model performance.

The comparative experimental results of difficulty distribution are shown in Experiment 1 in Table~\ref{tab:Difficulty levels}; and the comparison results of difficulty and length are presented in Experiment 2 in Table~\ref{tab:Difficulty levels}. In Table~\ref{tab:Scale and Curriculum Learning}, experiment 3 shows the comparison of data scale; and experiment 4 shows the comparison of curriculum learning.

\begin{table}[t]
\centering
\caption{Difficulty levels Experiment Results}
\label{tab:Difficulty levels}
\small
\begin{tabular}{lccccc}
\toprule
\multirow{2}{*}{Experiments} & \multicolumn{3}{c}{Experiment 1} & \multicolumn{2}{c}{Experiment 2} \\
\cmidrule(lr){2-4} \cmidrule(lr){5-6}
Base Model & \multicolumn{3}{c}{Qwen3-8B-base} & \multicolumn{2}{c}{Qwen3-8B-base} \\
\midrule
Data & 75K\_hard & 75K\_mid & 75K\_easy & 75K\_hard & 75K\_long \\
\midrule
IFEval & 26.99 & \textbf{28.10} & 24.03 & 26.99 & \textbf{32.35} \\
LiveCodeBench v5 & 12.87 & \textbf{13.47} & 11.38 & 12.87 & \textbf{18.26} \\
GPQA-Diamond & \textbf{47.47} & 42.42 & 44.95 & \textbf{47.47} & 45.96 \\
AIME'24 & \textbf{48.33} & 42.08 & 36.25 & 48.33 & \textbf{54.17} \\
AIME'25 & \textbf{33.75} & 29.17 & 31.25 & 33.75 & \textbf{37.50} \\
MATH-500 & 89.20 & 87.20 & \textbf{90.00} & 89.20 & \textbf{91.40} \\
\midrule
Average & 43.10 & 40.41 & 39.64 & 43.10 & \textbf{46.61} \\
\bottomrule
\end{tabular}
\end{table}

\begin{table}[t]
\centering
\small
\setlength{\tabcolsep}{3pt}
\caption{Scale and Curriculum Learning Experiment Results}
\label{tab:Scale and Curriculum Learning}
\begin{tabular}{lcccccc}
\toprule
\multirow{2}{*}{Experiments} & \multicolumn{3}{c}{Experiment 3} & \multicolumn{3}{c}{Experiment 4} \\
\cmidrule(lr){2-4} \cmidrule(lr){5-7}
Base Model & \multicolumn{3}{c}{Qwen3-8B-base} & \multicolumn{1}{c}{Qwen3-8B-base} & \multicolumn{2}{c}{75K\_short} \\
\midrule
Data & 50K\_long & 75K\_long & 150K\_all & 150K\_all & \multicolumn{2}{c}{75K\_long} \\
\midrule
IFEval & 25.88 & \textbf{32.35} & 26.80 & \textbf{26.80} & \multicolumn{2}{c}{22.92} \\
LiveCodeBench v5 & 10.78 & \textbf{18.26} & 10.48 & 10.48 & \multicolumn{2}{c}{\textbf{10.78}} \\
GPQA-Diamond & \textbf{48.48} & 45.96 & 47.47 & \textbf{47.47} & \multicolumn{2}{c}{46.46} \\
AIME'24 & 42.50 & 54.17 & 55.00 & 55.00 & \multicolumn{2}{c}{\textbf{57.08}} \\
AIME'25 & 31.67 & \textbf{37.50} & 37.50 & 37.50 & \multicolumn{2}{c}{\textbf{41.25}} \\
MATH-500 & 87.80 & \textbf{91.40} & 90.60 & 90.60 & \multicolumn{2}{c}{\textbf{91.40}} \\
\midrule
Average & 41.19 & \textbf{46.61} & 44.64 & 44.64 & \multicolumn{2}{c}{44.98} \\
\bottomrule
\end{tabular}
\end{table}

\textbf{Results and conclusions:} 
\begin{itemize}
    \item First, sample difficulty has a significant impact on complex mathematical reasoning ability. On AIME'24 and AIME'25, 75K\_hard outperforms 75K\_easy by 12.08 and 2.50 points, respectively, indicating that high-difficulty samples are more effective for improving the model's ability to solve complex reasoning problems. However, on MATH-500, 75K\_easy scores higher than 75K\_hard, suggesting that MATH-500 contains more basic or medium-difficulty problems and that simple samples are somewhat better matched to this benchmark. Therefore, relying solely on simple samples cannot bring the best overall gains; high-difficulty samples remain crucial for improving complex reasoning ability.

    \item Second, length-based sampling is superior to pure difficulty-based sampling in mathematical reasoning tasks. Compared with 75K\_hard, 75K\_long improves AIME'24, AIME'25, and MATH-500 by 5.84, 3.75, and 2.20 points. This indicates that, for Reasoning models, long chain-of-thought samples provide not only final answers but also more complete problem decomposition, derivation steps, and intermediate logic, thereby helping the model learn a more stable step-by-step reasoning pattern.

    \item Third, increasing data volume brings benefits, but the gains do not grow linearly with scale. When the data increases from 50K\_long to 75K\_long, the model improves significantly on AIME'24, AIME'25, and MATH-500, showing that sufficient high-quality reasoning samples are the foundation for forming reasoning ability. However, when the data is expanded to 150K\_all, AIME'24 improves by only 0.83 points, AIME'25 remains unchanged, and MATH-500 drops by 0.80 points. This suggests that if the added data contains many low-quality samples or weak reasoning chains, further scaling may dilute high-quality reasoning patterns and limit improvements in advanced reasoning ability.

    \item Finally, the two-stage short-to-long curriculum learning strategy outperforms single-stage mixed training. Under the same overall data range, training first on 75K\_short to establish a basic reasoning format and then on 75K\_long to strengthen long-chain reasoning improves AIME'24, AIME'25, and MATH-500 by 2.08, 3.75, and 0.80 points, respectively, compared with single-stage training on 150K\_all. This result indicates that curriculum learning allows the model to first acquire a basic problem-solving structure and then gradually adapt to longer and more complex reasoning processes, thereby shaping mathematical reasoning ability more stably.
\end{itemize}

\paragraph{Experiment on Data-Type Mixture Ratios}

\textbf{Data sources and composition:} After determining the filtering strategy for individual data categories, we further address the mixture ratio among multiple types of SFT corpora. In this experiment, we focus on three verifiable and reasoning-intensive domains: math, code, and science. These domains contribute differently to model capabilities: math data directly affects symbolic and multi-step reasoning ability; code data helps with structured problem solving and executable output formatting; and science data improves factual reasoning and scientific problem solving. Therefore, determining the mixture ratio among these data types is a key problem in SFT data construction.

\begin{figure}[t]
\centering
\includegraphics[width=\linewidth]{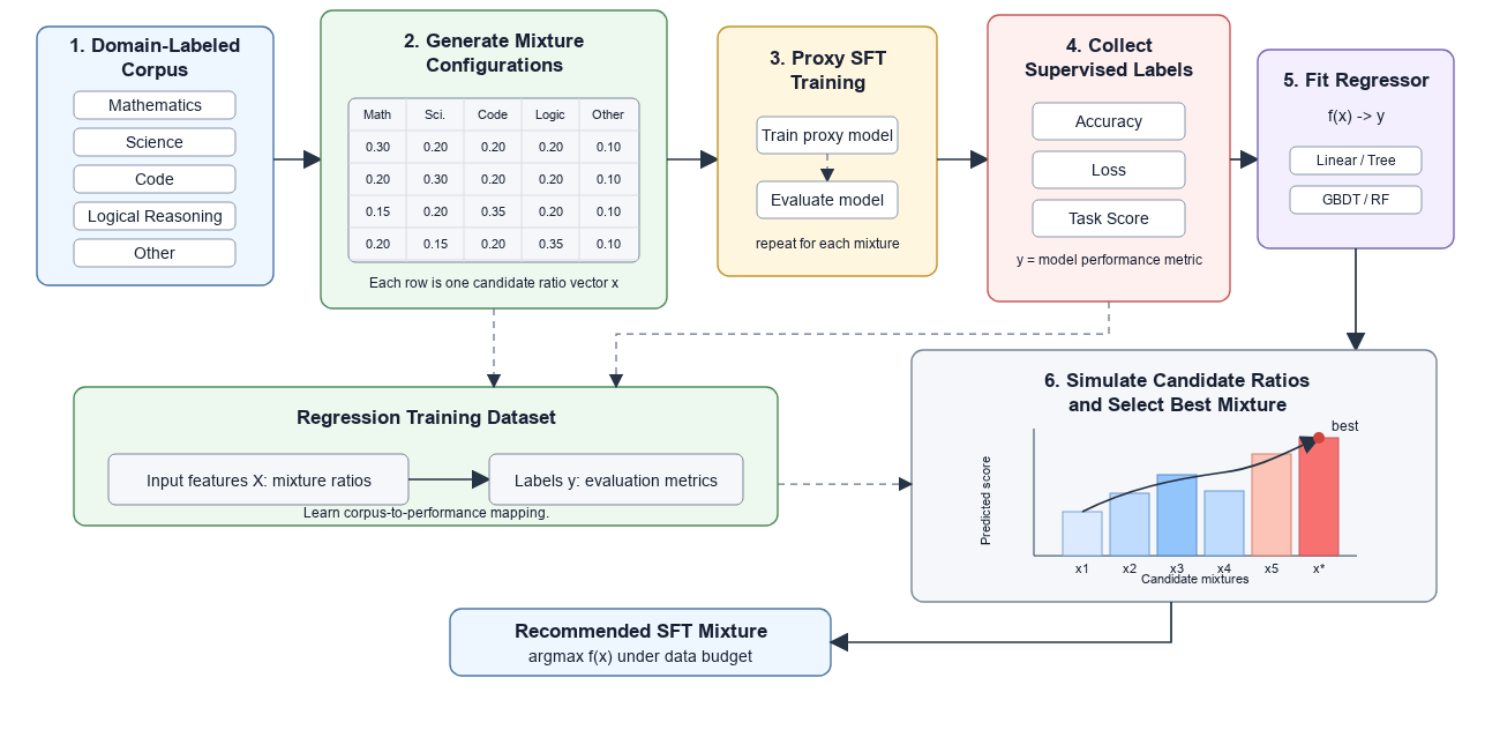}
\caption{Overview of the regression-based data mixing strategy. The process consists of three stages: (1) generating diverse mixture configurations from domain-labeled reasoning corpora, (2) conducting proxy training to collect performance labels, and (3) fitting a regression model to predict optimal mixture ratios for final SFT training.}
\label{fig:data_mixing_flow}
\end{figure}

\textbf{Data processing method:} As shown in Fig.~\ref{fig:data_mixing_flow}, we adopt a regression-based data mixing strategy, transforming the search for multi-domain data mixture ratios into a regression prediction problem. Specifically, multiple differentiated data mixture configurations are first generated according to the math, code, and science domain labels of the reasoning corpus. Proxy training is then conducted based on these configurations to obtain model evaluation results corresponding to each mixture. Finally, the data mixture ratios are used as input features, and the average benchmark gain is used as the supervised label to fit the mapping between corpus mixture ratios and model performance.

\textbf{Experimental design:} To reduce the cost of large-scale trial training, we do not directly conduct full training for all candidate mixtures. Instead, we first perform a cold-start SFT stage from Qwen3-8B-base to give the model basic reasoning ability. The subsequent mixture-ratio proxy experiments are then conducted on top of this cold-start checkpoint, so that the effect of different math, code, and science proportions can be evaluated as marginal changes in reasoning, coding, and scientific problem-solving metrics rather than as basic capability acquisition from scratch. We fit a lightweight regression model using the mixture ratios as input features and the average benchmark gain as the prediction target. At the same time, the cost of each trial is reduced by shortening training cycles and controlling sample scale. After the regression model is trained, we select candidate optimal mixtures suitable for Reasoning SFT according to the prediction results and use the selected mixture for subsequent full SFT training verification.

\begin{figure}[t]
\centering
\includegraphics[width=\linewidth]{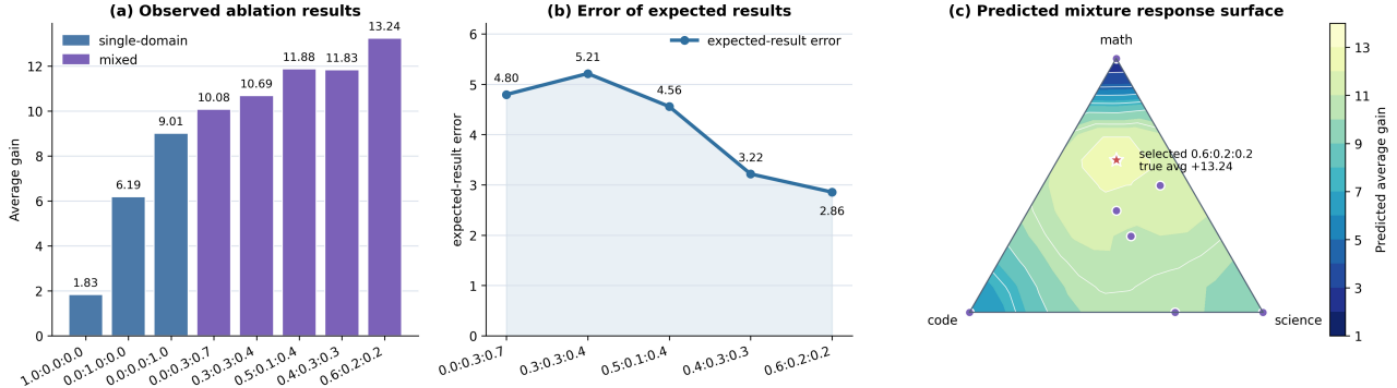}
\caption{Experimental results of the regression-based data mixture optimization strategy. 
(a) Observed performance gains of representative single-domain and mixed-data configurations. 
(b) Prediction error during the mixture search process: error is large in the early exploration stage but gradually becomes more stable as additional mixture experiments are incorporated. 
(c) Predicted performance surface over the mixture simplex; the star marks the selected optimal mixture.}
\label{fig:mixture_ratios}
\end{figure}

\textbf{Results and conclusions.}
Figure~\ref{fig:mixture_ratios} reports the empirical results of our regression-based data mixture optimization. 
Across all evaluated configurations, mixed-domain data consistently outperform single-domain ones, indicating complementary supervision from math, code, and science sources. 
The optimal mixture is 0.6:0.2:0.2 (math:code:science), which yields the highest average reasoning gain; other mixtures (e.g., 0.3:0.4:0.3, 0.5:0.4:0.1) also exceed single-domain baselines but remain suboptimal. 
This suggests that a math-dominant composition with moderate auxiliary data is necessary for balanced cross-task generalization.

An important observation from the early proxy experiments is that single-domain math training performed poorly on coding evaluation: its LiveCodeBench score was only 14.83. 
Manual inspection showed that the model often produced responses with malformed code formatting, making it difficult for the evaluator to extract executable code blocks. 
For this reason, the initial mixture search did not immediately increase the proportion of math data. 
However, once math data were combined with code data, the code-formatting issue was largely resolved, and the model began to benefit from the reasoning signal provided by math samples without losing code executability. 
As more mixture configurations were evaluated, the estimated contribution of math data became increasingly prominent, which explains why the final optimum shifts toward a math-dominant mixture while still retaining code data as a stabilizing component.

The key insight of our approach is to iteratively refine the mixture via a regression model that predicts performance from limited proxy experiments. 
As shown in Figure~\ref{fig:mixture_ratios}(b), the prediction error, measured by the absolute difference between predicted and observed average gains, is initially high due to sparse exploration, but decreases steadily as more mixture configurations are evaluated. 
Consequently, the observed empirical gains converge toward the predicted values, and the search progressively narrows down to the optimal region. 
Figure~\ref{fig:mixture_ratios}(c) visualizes the predicted response surface over the simplex; the best-performing area concentrates around math-dominant mixes rather than at vertices or extreme skews, confirming that data contributions are interdependent. 
The star marks the selected optimum, where the prediction aligns with empirical verification.

Together, these results validate the core idea: effective mixture search can be cast as a low-cost prediction problem. 
By sequentially incorporating proxy experiments, the regression model captures the composition–performance landscape and guides the selection of the final mixture. 
Based on both predicted surface and direct verification, we recommend the 0.6:0.2:0.2 ratio for full-scale SFT training.

\paragraph{Experiment on the Effectiveness of Data Filtering}

\textbf{Data sources and composition:} We conduct this experiment on the rejection-sampling-augmented data derived from the overall corpus described in Section~\ref{sec:data_curation}, with the goal of analyzing the impact of training data scale and data quality on model performance. We divide the experimental data into two training stages. The first-stage dataset establishes basic reasoning ability, while the original second-stage dataset is expanded through rejection sampling and contains more complex and diverse samples. Since the second-stage data are larger and come from more heterogeneous sources, data redundancy and noise are more likely to affect the effectiveness of continued training. Therefore, we focus on filtering and optimizing stage-2.

\textbf{Data processing method:} To improve the quality of stage-2 data, we select four core categories---math, code, science, and instruction following---and perform query-level deduplication and cleaning. After deduplication, the retained valid data sizes are 80 thousand code samples, 330 thousand mathematics samples, 190 thousand science samples, and 30 thousand instruction-following samples. To compress the data scale while preserving the original data distribution as much as possible, we remix the cleaned samples according to the category proportions in stage-2 and construct the stage-2-filter dataset. The final stage-2-filter dataset contains 240 thousand mathematics samples, 80 thousand code samples, 80 thousand science samples, and 30 thousand instruction-following samples.

\textbf{Experimental design:} All experiments use Qwen3-8B-base as the base model and include three comparison groups. The first group uses only stage-1 for first-stage training, in order to observe model performance after basic SFT. The second group continues training for one epoch on the original stage-2 data based on the stage-1 model. The third group continues training for five epochs on the filtered stage-2-filter data based on the stage-1 model. By comparing the second and third groups, this experiment evaluates the effectiveness of ``multi-epoch training on smaller high-quality data'' compared with ``single-epoch training on larger raw data.'' The final results are shown in Table~\ref{tab:Data Filtering}.

\begin{table}[t]
\centering
\caption{Data Filtering Experiment Results}
\label{tab:Data Filtering}
\small
\begin{tabular}{lrrr}
\toprule
Metric & stage-1 & stage-2 & stage-2-filter \\
\midrule
IFEval & 48.06 & 63.77 & \textbf{66.54} \\
LiveCodeBench v5 & 56.45 & 65.45 & \textbf{68.25} \\
GPQA-Diamond & 53.03 & \textbf{64.65} & 60.61 \\
AIME'24 & 58.75 & 77.92 & \textbf{80.83} \\
AIME'25 & 45.42 & 70.83 & \textbf{72.08} \\
MATH-500 & 94.80 & 97.40 & \textbf{98.00} \\
\bottomrule
\end{tabular}
\end{table}

\textbf{Results and conclusions:} 
\begin{itemize}
    \item In Table~\ref{tab:Data Filtering}, the results show that, compared with the original stage-2 data, stage-2-filter improves performance on IFEval, LiveCodeBench v5, AIME'24, AIME'25, and MATH-500 by 2.77, 2.80, 2.91, 1.25, and 0.60 points, respectively. This indicates that the combination of deduplication, quality filtering, and multi-epoch training on a compact dataset can effectively improve instruction following, code generation, and mathematical reasoning. The results suggest that data scale is not the only decisive factor in post-training. When the raw data contains redundancy and noise, a compact high-quality dataset trained for multiple epochs can often achieve higher training efficiency and better generalization than a larger raw dataset trained for a single epoch.

    \item We note that stage-2-filter decreases by 4.04 points on GPQA-Diamond compared with stage-2. This suggests that while the filtering strategy improves most capabilities, it may weaken coverage of some scientific question answering or high-difficulty knowledge samples. Therefore, subsequent data filtering should not only pursue deduplication and compression, but also preserve sufficient data diversity according to task type.
\end{itemize}
    
\subsubsection{SFT Model Results}

In the SFT stage, we use Qwen3-8B-base as the base model and train NebulaExp-reasoning-SFT with the high-quality dataset constructed through the corpus processing pipeline described in Section~\ref{sec:data_curation}. This stage focuses on multiple comparative experiments to analyze the impact of different data filtering strategies, data difficulty distributions, sample lengths, and data-type mixtures on model performance. Based on these experiments, the final SFT recipe is as follows: the optimal data-type mixture is 0.6:0.2:0.2 (math:code:science) determined via regression-based mixture search; long-chain reasoning samples are prioritized over pure difficulty-based selection; a two-stage short-to-long curriculum learning strategy is employed, where the model is first trained on shorter reasoning chains to establish basic reasoning formats and then on long-chain samples to strengthen multi-step reasoning; and the stage-2 dataset is deduplicated, quality-filtered, and trained for multiple epochs on a compact high-quality subset, improving training efficiency while reducing the negative impact of noisy or duplicated samples. 

We compare NebulaExp-reasoning-SFT against other models; the SFT evaluation results are shown in Table~\ref{tab:Evaluation Results}.

\begin{table}[t]
\centering
\small
\caption{SFT Evaluation Results}
\label{tab:Evaluation Results}
\begin{tabular}{lrrrr}
\toprule 
Metric & Qwen3-8B-thinking & Qwen3-14B & QWQ-32B &  NebulaExp-reasoning-SFT \\
\midrule
IFEval & 85.00 & \textbf{85.40} & 83.90 & 66.54 \\
LiveCodeBench v5 & 65.78 & 63.50 & 62.70 & \textbf{68.25} \\
GPQA-Diamond & 59.60 & 64.00 & \textbf{65.60} & 60.61 \\
AIME'24 & 76.00 & 79.30 & 79.50 & \textbf{80.83} \\
AIME'25 & 65.71 & 70.40 & 69.50 & \textbf{72.08} \\
MATH-500 & 96.60 & 96.80 & \textbf{98.00} & \textbf{98.00} \\
Average & 74.78 & \textbf{76.57} & 76.53 & 74.39 \\
\bottomrule
\end{tabular}
\end{table}

\textbf{Results and conclusions:} 
\begin{itemize}
    \item Compared with the three baseline variants, NebulaExp-reasoning-SFT achieves an overall average (74.39) marginally below Qwen3-8B-thinking (74.78) and slightly inferior to the other two baselines. This performance discrepancy mainly stems from the deficit on IFEval and GPQA-Diamond. Since this stage uses only SFT optimization, the underwhelming IFEval result is expected. We anticipate that the performance gap will be largely narrowed in the subsequent RL training phase.
    
    \item On the code generation benchmark LiveCodeBench v5, NebulaExp-reasoning-SFT achieves an accuracy of 68.25, outperforming Qwen3-8B-thinking, Qwen3-14B, and QWQ-32B by 2.47, 4.75, and 5.55 percentage points, respectively. These consistent margins demonstrate that our data filtering pipeline strengthens the model's code generation capability.

    \item On the mathematical reasoning benchmarks AIME'24, AIME'25, and MATH-500, NebulaExp-reasoning-SFT consistently surpasses all three baselines. On AIME'24, NebulaExp-reasoning-SFT attains an accuracy of 80.83, exceeding Qwen3-8B-thinking, Qwen3-14B, and QWQ-32B by 4.83, 1.53, and 1.33 percentage points, respectively. On AIME'25, it further reaches 72.08, yielding clear improvements of 6.37, 1.68, and 2.58 percentage points over the same three baselines. On MATH-500, our method achieves 98.00, matching the strongest baseline QWQ-32B while surpassing Qwen3-8B-thinking and Qwen3-14B by 1.40 and 1.20 percentage points, respectively. Taken together, these consistent gains across all three benchmarks confirm the effectiveness of our sampling strategy for constructing high-quality mathematical reasoning corpora.
\end{itemize}

\subsection{Reinforcement Learning for Reasoning Models}

\subsubsection{RL Technical Route and Details}
To address systematic deficiencies in specific reasoning patterns for complex tasks such as mathematics, code generation, and scientific problem-solving, we identify poorly performing problem categories from the reasoning validation set and construct additional training examples or design targeted reward shaping for these weak domains. Standard SFT often achieves high overall accuracy but fails to resolve long-tailed weaknesses, and its performance gain diminishes rapidly after a certain point. By introducing RL and focusing parameter updates on these identified weak spots, we effectively improve performance on the most difficult reasoning subsets. This targeted strategy avoids inefficient uniform optimization and directly enhances the model’s lower-bound performance in complex reasoning tasks.

The Reasoning RL stage also inherits the empirical lessons from the Instruct RL experiments above. Specifically, we adopt the best-performing practical recipe identified there: training on verifiable samples rather than the full raw pool, filtering out samples that are either already mastered or provide no useful advantage signal, retaining KL regularization for stable optimization, and using GRPO with correctness- and format-based rewards. In this section, ``medium-difficulty'' is defined relative to NebulaExp-reasoning-SFT and denotes samples that remain learnable while still providing non-trivial advantage signals; it is therefore not identical to the hard-only split used in the Instruct RL ablation. This design keeps the Reasoning RL experiment consistent with the validated Instruct RL setup while adapting the data mixture to reasoning-oriented weaknesses.

\subsubsection{RL Training Results and Key Findings}
We source Reasoning RL data from the 200K RL candidate pool, with the final samples drawn from the Nemotron-Cascade and Olmo3 collections. Using NebulaExp-reasoning-SFT as an evaluator, we classify each sample into three difficulty levels (easy, medium, hard) based on prediction correctness and confidence. Only medium-difficulty samples are retained for the actual Reasoning RL training subset, yielding a balanced set of 8,000 examples: 4,000 instruction-following, 2,000 mathematical, 1,000 scientific, and 1,000 code generation instances.
The base model for this stage is NebulaExp-reasoning-SFT. We train it with GRPO, using rewards defined solely by answer correctness and format compliance (no auxiliary process reward model). Hyperparameters are listed in Table~\ref{tab:hyper}. 
\begin{table*}[t]
\centering
\caption{Hyperparameters for reinforcement learning.}
\label{tab:hyper}
\small
\begin{tabular}{lc}
\toprule
Hyper-parameter          & Value \\
\midrule
Maximum response length  & 32K   \\
Batch size               & 64    \\
Rollout size             & 8     \\
Learning rate            & $2\times10^{-6}$ \\
Optimizer                & AdamW ($\beta_1=0.9$, $\beta_2=0.95$) \\
Temperature              & 1.0   \\
Top-$p$                  & 0.95  \\
Overlong filtering       & False \\
\bottomrule
\end{tabular}
\end{table*}

\begin{figure*}[t]
\centering
\includegraphics[width=\textwidth]{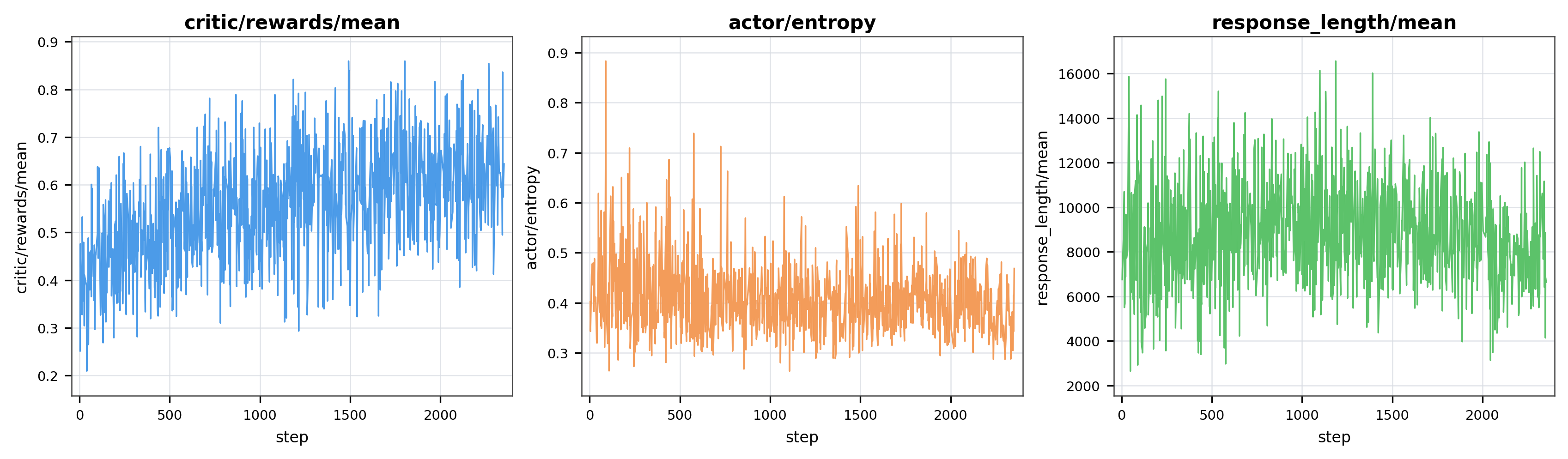}
\caption{Training curves of the final Reasoning RL model on the 8K training subset. The three panels show (a) critic/rewards/mean, (b) actor/entropy, and (c) response\_length/mean over the full training run. }
\label{fig:reasoning_rl_training_curves}
\end{figure*}

As shown in Figure~\ref{fig:reasoning_rl_training_curves}, the training process converges smoothly: the mean reward steadily increases and plateaus around step~1,500, while the actor entropy gradually declines, indicating reduced policy uncertainty. Concurrently, the response length grows and stabilizes, suggesting that the model learns to generate longer reasoning chains when beneficial. These trends are consistent with the performance improvements observed in Table~\ref{tab:results}, particularly for epoch~2.

\begin{table*}[t]
\centering
\caption{Performance on reasoning benchmarks (excluding IFEval). Parentheses show the difference from NebulaExp-reasoning-SFT. Bold numbers indicate the best performance per dataset (including the external baseline).}
\label{tab:results}
\small
\begin{tabular}{lcccccc}
\toprule
Dataset & Qwen3-8B-thinking & NebulaExp-reasoning-SFT & RL-epoch1 & RL-epoch2 & RL-epoch3 \\
\midrule
LiveCodeBench v5  & 65.78 & 65.17 & 65.50\,(+0.33)  & 65.73\,(+0.56)  & \textbf{68.86}\,(+3.69)  \\
GPQA-Diamond      & 59.60 & 58.08 & 60.10\,(+2.02)  & \textbf{62.63}\,(+4.55)  & 61.11\,(+3.03)  \\
AIME'24          & 74.29 & 77.92 & \textbf{79.17}\,(+1.25)  & \textbf{79.17}\,(+1.25)  & 76.67\,(-1.25)  \\
AIME'25          & 65.71 & 70.83 & 68.75\,(-2.08)  & \textbf{72.50}\,(+1.67)  & 72.08\,(+1.25)  \\
MATH-500          & 96.60 & \textbf{97.40} & 97.00\,(-0.40)  & 95.80\,(-1.60)  & 97.00\,(-0.40)  \\
\midrule
Average           & 72.40 & 73.88 & 74.10 & \textbf{75.17} & 75.14 \\
\bottomrule
\end{tabular}
\end{table*}

 RL training on medium-difficulty data improves the average reasoning benchmark score, with epoch~2 achieving the best overall average. Coding performance on LiveCodeBench v5 improves only modestly after the first two epochs---$0.33$ and $0.56$ percentage points respectively---but then jumps to $3.69$ percentage points at epoch~3, surpassing the Qwen3-8B-thinking baseline. Scientific reasoning on GPQA-Diamond improves consistently, reaching a maximum gain of $4.55$ percentage points at epoch~2. Mathematical results are mixed. Epoch~2 yields the best gains on AIME'24 and AIME'25, with increases of $1.25$ and $1.67$ percentage points respectively, but epoch~3 regresses on AIME'24 by $1.25$ points. On the nearly saturated MATH-500 benchmark, where NebulaExp-reasoning-SFT already achieves $97.4\%$, all RL versions underperform slightly by $0.4$ to $1.6$ percentage points, likely due to exploration noise. Overall, epoch~2 provides the best average and the strongest trade-off across the five reported reasoning benchmarks; epoch~3 improves coding while regressing on several reasoning tasks.

Compared with the external baseline Qwen3-8B-thinking, NebulaExp-reasoning-SFT already leads on AIME and MATH-500 but lags on GPQA-Diamond. After two RL epochs, the model surpasses the baseline on GPQA-Diamond with $62.63$ versus $59.60$. These results demonstrate that RL on a curated 8K subset of medium-difficulty data can effectively enhance weak reasoning in an 8B parameter model, with the largest gains in science and selected reasoning tasks, while saturated math benchmarks may see slight regressions. Based on these findings, two RL epochs are recommended.

\subsection{On-Policy Distillation for Reasoning Model}

Our preceding RL experiments (\S\ref{sec:reasoning_training}) demonstrate that reinforcement learning with correctness-based reward signals can effectively improve reasoning models on verifiable tasks. However, RL methods face several inherent limitations in practice. First, a broad range of real-world tasks---such as creative writing, text summarization, and open-domain question answering---lack reliable verifiers and cannot provide effective reward signals. Second, the sparse rewards and high-variance gradients inherent to RL training make the process sensitive to hyperparameters, with convergence stability inferior to supervised fine-tuning. Third, optimizing for a specific capability via RL often comes at the expense of other capabilities. We therefore explore alternative or complementary post-training optimization methods beyond RL. On-Policy Distillation (OPD), an optimization pathway that does not depend on reward signals, has recently been successfully applied in models such as DeepSeek-V4~\cite{deepseek_v4} and GLM-5~\cite{zeng2026glm}. We investigate the effectiveness of OPD in reasoning model post-training through two experiments: single-teacher OPD for instruction following and multi-teacher OPD for multi-capability distillation. The central question is whether OPD can deliver further improvements compared to RL.

\subsubsection{OPD Technical Route and Details}

\paragraph{Comparison of RL and OPD.}
RL and OPD share the same ultimate goal---further improving model capability on top of an SFT base---but differ fundamentally in their optimization pathways. RL drives policy updates through trial-and-error exploration guided by reward signals, with the optimization direction defined by task-specific verifier functions. OPD, in contrast, transfers knowledge by imitating the output distribution of a teacher model, with the optimization direction defined by the teacher's behavioral distribution. Table~\ref{tab:rl_vs_opd} compares the two approaches across several dimensions.

\begin{table}[t]
\centering
\caption{Comparison of RL and OPD.}
\label{tab:rl_vs_opd}
\small
\setlength{\tabcolsep}{6pt}
\begin{tabular}{@{}lcc@{}}
\toprule
\textbf{Dimension} & \textbf{RL} & \textbf{OPD} \\
\midrule
Optimization signal  & Reward function (sparse)  & Teacher distribution (dense, per-token) \\
Requires verifier    & Yes                       & No \\
Training stability   & Lower (high-variance gradients) & Higher (low-variance KL gradients) \\
Multi-capability coverage & Multi-reward weighting & Multi-teacher distillation \\
Computational cost   & High (online sampling + reward computation) & Moderate (teacher forward pass) \\
Task generalization  & Verifiable tasks only     & Any task type \\
Optimization ceiling & Determined by reward design & Determined by teacher capability \\
\bottomrule
\end{tabular}
\end{table}

\paragraph{Core Principles of OPD.}
Conventional Knowledge Distillation (KD) trains the student model on teacher-generated data~\cite{hinton2015distilling}. For autoregressive language models, this means first sampling an output sequence $y$ from the teacher distribution $\pi_T(\cdot|x)$, and then aligning the student distribution $\pi_S(\cdot|x)$ with the teacher via cross-entropy or KL divergence. However, this off-policy approach suffers from a severe distributional mismatch: during training, the student sees input contexts drawn from the teacher distribution, whereas at inference time the input contexts come from its own distribution. The resulting exposure bias limits the effectiveness of distillation~\cite{agarwal2023gkd}.

The core innovation of OPD is to have the student model generate training sequences from its own distribution, and then have the teacher provide per-token probability distribution supervision over these student-generated sequences. Concretely, given an input question $x$, the student model $\pi_S$ first autoregressively generates an output sequence $y \sim \pi_S(\cdot|x)$; the teacher model $\pi_T$ then computes conditional probability distributions $\pi_T(\cdot|y_{<t}, x)$ at each position $t$ of $y$; the student is then optimized to match these teacher distributions by minimizing the reverse KL divergence:

\begin{equation}
\mathcal{L}_{\text{OPD}} = \mathbb{E}_{x \sim \mathcal{D},\, y \sim \pi_S(\cdot|x)} \left[ \sum_{t=1}^{|y|} D_{\text{KL}}\Bigl(\pi_S(\cdot|y_{<t}, x) \;\Big\|\; \pi_T(\cdot|y_{<t}, x)\Bigr) \right]
\label{eq:opd_loss}
\end{equation}

where $\mathcal{D}$ is the set of training questions. There are two key differences from standard KD: first, the sampling source $y \sim \pi_S(\cdot|x)$ comes from the student itself (on-policy), eliminating the train--inference distribution mismatch; second, the use of reverse KL (with $\pi_S$ in the first argument and $\pi_T$ in the second) drives the student distribution to actively move toward the teacher distribution, yielding more stable gradients.

\paragraph{Implementation Details.}
In our engineering implementation, we adopt the following configuration to balance training efficiency and distillation quality. During generation, we use a temperature of $T=1.0$ to preserve the student model's ability to explore fully. When computing the KL divergence, to avoid the prohibitive memory cost of full-vocabulary computation (typically 128K+ tokens), we restrict the calculation to the top-64 tokens with the highest probability in the teacher distribution, treating the probability mass of all remaining tokens as zero and ignoring them. The teacher model is frozen during training (teacher-freeze) and only provides logits as supervision targets. The entire training process is fully on-policy, with no teacher-forcing mixing strategy.

\paragraph{Training Hyperparameters.}
OPD training uses the following fixed hyperparameters: learning rate $5 \times 10^{-6}$, batch size 32, AdamW optimizer ($\beta_1=0.9$, $\beta_2=0.95$), cosine learning rate decay, and a maximum sequence length of 32K tokens. Training proceeds for 3 epochs, with checkpoints saved and evaluated at fixed step intervals. All single-teacher OPD experiments are implemented using the ms-swift framework with ZeRO-2 optimization. For multi-teacher OPD, we use the verl framework with learning rate $1\times10^{-6}$, batch size 32, 1 epoch, and gradient clipping at $[-3, 3]$.

\paragraph{Multi-Teacher OPD.}
While a single teacher model may be broadly capable, it is not necessarily optimal on every specific capability dimension. Mirroring the RL paradigm where independent rewards can be designed for different capability domains, OPD naturally supports assigning domain-specialist teacher models to different capability domains. Formally, for each capability domain $d \in \mathcal{D} = \{\text{math}, \text{code}, \text{science}, \text{general}\}$, we assign a teacher $\pi_T^{(d)}$ that performs best in that domain, and route the distillation loss by sample domain:

\begin{equation}
\mathcal{L}_{\text{MT-OPD}} = \sum_{d \in \mathcal{D}} \alpha_d \cdot \mathbb{E}_{\substack{x \sim \mathcal{D}_d \\ y \sim \pi_S(\cdot|x)}} \left[ \sum_{t=1}^{|y|} D_{\text{KL}}\Bigl(\pi_S(\cdot|y_{<t}, x) \;\Big\|\; \pi_T^{(d)}(\cdot|y_{<t}, x)\Bigr) \right]
\label{eq:mt_opd_loss}
\end{equation}

where $\mathcal{D}_d$ is the training subset for domain $d$, and $\alpha_d$ are domain weights ($\sum_d \alpha_d = 1$) set proportionally to the amount of training data in each domain. The key advantage of multi-teacher OPD is that the distillation signal for each capability domain comes from the strongest teacher in that domain, avoiding the quality degradation that occurs when a single teacher underperforms on a specific domain.

\subsubsection{OPD Training Results and Key Findings}

To systematically validate the effectiveness of OPD, we conduct two sets of experiments. The first focuses on single-teacher OPD with instruction following (IF) as the target capability, comparing different teacher models and directly addressing whether OPD can surpass RL. The second extends to multi-teacher OPD (MOPD), using domain-specialist 8B models obtained through RL training as teachers to verify the potential for multi-capability fusion.

\paragraph{Single-Teacher OPD for Instruction Following.}

\textbf{Experimental Setup.}
This experiment takes instruction following as the optimization target. IF was chosen for two reasons: (1) instruction following is a known weakness of current reasoning models---as shown in the preceding RL experiments, the SFT baseline achieves an IFEval score of only 69.00 (thinking mode), substantially lower than other capability dimensions; (2) IF tasks involve diverse and complex evaluation criteria (multiple constraints, formatting requirements, etc.) that are difficult to capture precisely with a single reward function, making them an ideal scenario for validating OPD's verifier-free advantage.

The base model is the same SFT checkpoint used in the preceding RL training, which already possesses basic reasoning and instruction-following capabilities. Training data are extracted from the RL training corpus, retaining only IF-type samples, yielding 3,960 training samples and 40 validation samples, totaling approximately 4K. To systematically compare the effect of different teacher models on OPD, we select four models from the Qwen3 family with varying sizes and training maturity: (1) Qwen3-8B---same size and vocabulary as the base model, serving as a same-family control; (2) Qwen3-235B-A22B---a large-parameter MoE model, to examine the effect of teacher capacity on distillation; (3) Qwen3-32B---a large-parameter Dense model, for comparison with the MoE variant; (4) Qwen3-30B-A3B and Qwen3-30B-A3B-Thinking-2507, to examine the effect of teacher training maturity.

\textbf{Evaluation Protocol.}
Since this experiment focuses on IF capability, the primary evaluation metric is IFEval (strict-prompt accuracy, thinking mode). In addition, to observe the indirect effects of OPD on general capabilities, we select the best IF checkpoint from each group for general capability evaluation, with the benchmark suite comprising: GPQA-Diamond (scientific reasoning), AIME'25 (math competition), MATH-500 (math fundamentals), C-Eval (Chinese general knowledge), and MMLU-Redux (English general knowledge).

\textbf{IF Capability Improvement Results.}
Table~\ref{tab:opd_if_progression} presents the IFEval scores under different teacher models.

\begin{table*}[t]
\centering
\caption{IFEval scores under different teacher models (thinking mode). Base model IFEval = 69.00, RL baseline = 80.14. Best values in bold; values in parentheses indicate improvement over the base model.}
\label{tab:opd_if_progression}
\small
\setlength{\tabcolsep}{4.5pt}
\begin{tabular}{@{}lcccccc@{}}
\toprule
\textbf{Teacher Model} & \textbf{Teacher IFEval} & \textbf{Max IFEval} & \textbf{$\Delta$ vs.\ Base} & \textbf{$\Delta$ vs.\ RL} & \textbf{Epochs} & \textbf{Learning Rate} \\
\midrule
Qwen3-8B                & 85.40 & 77.83 & $+$8.83  & $-$2.31 & 1 & $3\times10^{-6}$ \\
Qwen3-8B                & 85.40 & \textbf{82.24} & \textbf{$+$13.24} & \textbf{$+$2.10} & 3 & $5\times10^{-6}$ \\
\midrule
Qwen3-235B-A22B         & 84.45 & 74.81 & $+$5.81  & $-$5.33 & 3 & $5\times10^{-6}$ \\
Qwen3-32B               & 85.97 & 77.45 & $+$8.45  & $-$2.69 & 3 & $5\times10^{-6}$ \\
Qwen3-30B-A3B           & 86.37 & 80.55 & $+$11.55 & $+$0.41  & 3 & $5\times10^{-6}$ \\
Qwen3-30B-A3B-Thinking-2507 & 87.75 & \textbf{83.40} & \textbf{$+$14.40} & \textbf{$+$3.26} & 2 & $5\times10^{-6}$ \\
\bottomrule
\end{tabular}
\end{table*}

\textbf{General Capability Evaluation.}
From the above experimental groups, we select the two configurations with the best IFEval scores---OPD-T8B (teacher: Qwen3-8B, IFEval = 82.24) and OPD-T30B (teacher: Qwen3-30B-A3B-Thinking-2507, IFEval = 83.40)---for general capability evaluation, comparing them against the RL baseline and their respective teacher models. Results are shown in Table~\ref{tab:opd_general}.

\begin{table*}[t]
\centering
\caption{General capability evaluation of OPD models (thinking mode). AVER denotes the arithmetic mean.}
\label{tab:opd_general}
\small
\setlength{\tabcolsep}{4.5pt}
\begin{tabular}{@{}lccccccc@{}}
\toprule
\textbf{Model} & \textbf{GPQA} & \textbf{AIME'25} & \textbf{MATH-500} & \textbf{C-Eval} & \textbf{MMLU-Redux} & \textbf{IFEval} & \textbf{AVER} \\
\midrule
Base (SFT ckpt) & 57.58 & 59.04 & 96.80 & 31.12 & 69.75 & 69.00 & 63.88 \\
RL              & 57.58 & 62.37 & 96.40 & 49.79 & 74.82 & 80.14 & 70.18 \\
\midrule
OPD-T8B         & 54.05 & 69.04 & 96.00 & 59.90 & 84.83 & 82.24 & 74.34 \\
OPD-T30B        & 53.04 & 62.37 & 97.20 & 66.73 & 84.89 & 83.40 & 74.61 \\
\midrule
Teacher-8B      & 59.60 & 65.71 & 96.60 & 81.73 & 85.86 & 85.40 & 79.15 \\
Teacher-30B     & \textbf{66.67} & \textbf{72.37} & \textbf{97.80} & \textbf{83.28} & \textbf{90.15} & \textbf{87.75} & \textbf{83.00} \\
\bottomrule
\end{tabular}
\end{table*}

\textbf{Analysis.}
As shown in Table~\ref{tab:opd_if_progression}, with a suitable teacher model and hyperparameter configuration, OPD can surpass the RL baseline in IF capability. Using Qwen3-8B as teacher (same size, same vocabulary), IFEval reaches 82.24, exceeding the RL baseline (80.14) by 2.10 points. Using Qwen3-30B-A3B-Thinking-2507 as teacher, IFEval further improves to 83.40, leading RL by 3.26 points. In terms of general capability (Table~\ref{tab:opd_general}), OPD-T8B and OPD-T30B achieve average scores of 74.34 and 74.61, respectively, both surpassing RL's 70.18. Notably, OPD training used only 4K IF samples, far fewer than the 53K samples used in RL training, and the training process was more stable (free from sparse reward and variance issues), highlighting OPD's advantages in data efficiency and training stability.

The teacher model comparison reveals three important patterns. First, teachers from the same base model family perform best: Qwen3-8B as teacher yields an IFEval gain of +13.24, outperforming the larger Qwen3-235B-A22B (+5.81) and Qwen3-32B (+8.45). This is because the 8B teacher shares the same vocabulary and a similar output distribution space with the base model, resulting in better-aligned distillation signals. This finding is consistent with prevailing industry practice---training different capability branches from a shared base model and then fusing them via OPD into a unified model~\cite{deepseek_v4}. Second, MoE vs.\ Dense architecture does not affect OPD: Qwen3-30B-A3B (2507 version), also an MoE architecture, achieves the best result (+14.40), while 235B-A22B (likewise MoE) performs the worst (+5.81), indicating that architecture is not the primary factor. Third, teacher capability matters more than teacher size: the teacher model's score on the target capability is positively correlated with the OPD improvement magnitude (the IF capability of 235B-A22B is in fact lower than that of 8B), rather than being a simple monotonic function of parameter count. Therefore, teacher selection for OPD should prioritize target capability over model scale.

As can be seen from Table~\ref{tab:opd_general}, although OPD training used only IF data, the model exhibits substantial improvements on non-IF benchmarks such as AIME'25, MATH-500, C-Eval, and MMLU-Redux (e.g., C-Eval improved from 31.12 to 66.73, MMLU-Redux from 69.75 to 84.89). This cross-capability transfer may stem from two sources. First, the enhanced IF capability improves the model's overall instruction adherence, indirectly raising output quality across all benchmarks. Second, during white-box distillation, even though the training data are IF scenarios, the teacher model's output logits encode rich world knowledge and reasoning patterns, which the student model indirectly acquires in the process of approximating the teacher distribution. The sole exception is GPQA-Diamond, which declines after OPD (57.58 $\to$ 53.04), possibly related to the model generating more concise responses after OPD training; this could be remedied by incorporating scientific reasoning data in future work.

A gap of approximately 3--8 points remains between OPD models and their teachers (e.g., OPD-T8B IFEval = 82.24 vs.\ teacher = 85.40, AVER = 74.34 vs.\ 79.15; OPD-T30B IFEval = 83.40 vs.\ teacher-30B = 87.75, AVER = 74.61 vs.\ 83.00). This gap indicates that the absolute capability of the teacher constitutes a ceiling for OPD performance. Future optimization directions include using stronger teacher models, extending the number of training epochs, and introducing curriculum learning strategies.

\paragraph{Multi-Teacher OPD for Multi-Capability Distillation.}

\textbf{Objective.}
The single-teacher experiment verified that OPD can surpass RL in the IF scenario. However, a single teacher model is typically not optimal across all capability dimensions simultaneously. This experiment employs Multi-Teacher OPD (MOPD), using domain-specialist 8B models obtained through RL training as teachers, to verify whether fusing multiple specialist teachers through OPD can simultaneously outperform the base model and external baselines across multiple capability dimensions.

\textbf{Teacher Model Training and Selection.}
The MOPD experiment proceeds in two stages. In the first stage, domain-specialist RL teacher models are trained separately for four capability domains: math, science, code, and instruction following. For each domain, 2K training samples are randomly drawn from the medium-difficulty subset of the RL data pool. The math, science, and code teachers use SFT domain-expert models as their base, while the IF teacher uses Qwen3-8B as its base. All teachers are trained using the DAPO algorithm with GRPO advantage estimation, learning rate $1\times10^{-6}$, batch size 64, 4 epochs, rollout size 4, implemented via the verl framework.

The evaluation results of the four domain teacher models across training epochs are presented in Table~\ref{tab:mopd_teacher_selection}. Teacher model selection is based on best performance on each domain's core benchmark: the math domain selects epoch3 (AIME'25 = 75.42), and the science domain selects epoch3 (GPQA-Diamond = 62.12). The code and IF domains exhibit a noteworthy phenomenon: the model trained in the IF domain (IF-epoch3) achieves 73.74 on LiveCodeBench v5, exceeding the best result of the code-specialist model (epoch3, 71.33), while IF-epoch4 achieves 87.80 on IFEval. Consequently, the final MOPD teacher set consists of four checkpoints: Math-epoch3, Science-epoch3, IF-epoch3 (as the code teacher), and IF-epoch4 (as the IF teacher). This finding reveals an asymmetry in cross-domain capability transfer during RL training: the instruction comprehension and format control abilities acquired during IF training confer significant indirect benefits to code generation, whereas the transfer effects of code-specialist training to other domains are relatively limited.

\begin{table*}[t]
\centering
\caption{Evaluation results and final selection of the four domain RL teacher models (thinking mode). Best values in each domain are shown in bold; checkpoints selected as MOPD teachers are highlighted with a gray background. The baseline refers to the starting model for RL training in each domain. The IF teacher uses Qwen3-8B as its base; the other three domains use SFT domain-expert models.}
\label{tab:mopd_teacher_selection}
\small
\setlength{\tabcolsep}{3.5pt}
\begin{tabular}{@{}lccccccc@{}}
\toprule
\textbf{Domain} & \textbf{Checkpoint} & \textbf{GPQA} & \textbf{AIME'25} & \textbf{MATH-500} & \textbf{LCB v5} & \textbf{IFEval} \\
\midrule
\multirow{5}{*}{Math}
& baseline & 55.56 & 71.25 & 96.40 & 14.83 & -- \\
& epoch1   & 57.07 & 70.83 & 97.12 & 14.74 & 40.67 \\
& epoch2   & 56.06 & 72.92 & 97.15 & 15.21 & 42.51 \\
& \cellcolor{gray!25}\textbf{epoch3} & 52.53 & \textbf{75.42} & 96.97 & 16.07 & 41.96 \\
& epoch4   & 53.03 & 71.25 & \textbf{97.20} & 16.92 & 42.33 \\
\midrule
\multirow{5}{*}{Science}
& baseline & 61.11 & 63.33 & 96.40 & 53.08 & -- \\
& epoch1   & 61.11 & 70.83 & 96.65 & 55.92 & 42.33 \\
& epoch2   & 59.60 & 67.08 & 96.62 & --     & 44.92 \\
& \cellcolor{gray!25}\textbf{epoch3} & \textbf{62.12} & 68.75 & 96.60 & 57.20 & 43.07 \\
& epoch4   & 57.07 & 67.08 & \textbf{96.70} & --     & 41.96 \\
\midrule
\multirow{5}{*}{Code}
& baseline & 55.05 & 52.50 & 93.80 & 68.48 & -- \\
& epoch1   & 59.60 & 50.00 & 94.23 & 69.91 & 35.86 \\
& epoch2   & 54.55 & 48.33 & 94.58 & 69.57 & 35.30 \\
& epoch3   & 52.02 & 49.58 & 94.23 & 71.33 & 37.52 \\
& epoch4   & 50.00 & 50.83 & 93.92 & 69.95 & 36.41 \\
\midrule
\multirow{5}{*}{IF}
& baseline & 59.60 & 65.71 & 96.60 & 65.78 & 85.40 \\
& epoch1   & 56.57 & 67.92 & 96.53 & 72.84 & 85.95 \\
& epoch2   & 56.06 & 68.33 & 96.80 & 73.84 & 85.21 \\
& \cellcolor{gray!25}\textbf{epoch3} & 59.09 & 70.83 & 96.58 & \textbf{73.74} & 87.43 \\
& \cellcolor{gray!25}\textbf{epoch4} & \textbf{65.15} & 69.58 & 96.33 & 73.51 & \textbf{87.80} \\
\bottomrule
\end{tabular}
\end{table*}

\textbf{MOPD Training Setup.}
In the second stage, using the same mid-training SFT checkpoint as the base model (identical to the single-teacher experiment), MOPD training is conducted with the four selected teacher models. Training data are drawn from the same source as the teacher RL data but are independently sampled, totaling 10K samples: 2K math, 2K science, 2K code, and 4K IF. The domain proportions reflect the foundational role of IF data in multi-capability fusion. During student generation, we adopt a sampling strategy of temperature $T=0.6$ and top-$p=0.95$ to balance generation quality and diversity. KL loss computation uses top-64 truncation (loss mode: k1). The domain weights $\alpha_d$ of the teacher models are set proportionally to the amount of training data in each domain.

\textbf{Experimental Results.}
Table~\ref{tab:mopd_results} presents the MOPD training results. The checkpoint with the highest average score during training (epoch 2/3) is selected as the final MOPD model. The teacher best scores (the maximum across all four domain teachers on each benchmark, not necessarily from the same model) are also reported as an upper-bound reference for teacher capability.

\begin{table*}[t]
\centering
\caption{Multi-Teacher OPD experimental results (thinking mode). Values in parentheses indicate changes relative to Base. Teacher Best is the maximum score across all four domain teachers on each benchmark (not limited to a single model).}
\label{tab:mopd_results}
\small
\setlength{\tabcolsep}{5pt}
\begin{tabular}{@{}lcccc@{}}
\toprule
\textbf{Benchmark} & \textbf{Qwen3-8B} & \textbf{Base (SFT)} & \textbf{MOPD} & \textbf{Teacher Best} \\
\midrule
\multicolumn{5}{c}{\textit{Instruction Following}} \\
\midrule
IFEval          & 85.40 & 63.77 & 77.26\,($+$13.49) & 84.80 \\
\midrule
\multicolumn{5}{c}{\textit{Coding}} \\
\midrule
LiveCodeBench v5 & 65.78 & 65.45 & 69.53\,($+$4.08) & 73.74 \\
\midrule
\multicolumn{5}{c}{\textit{Scientific Reasoning}} \\
\midrule
GPQA-Diamond    & 59.60 & 64.65 & 63.13\,($-$1.52) & 65.15 \\
\midrule
\multicolumn{5}{c}{\textit{Mathematical Reasoning}} \\
\midrule
AIME'25         & 65.71 & 70.83 & 75.83\,($+$5.00) & 75.42 \\
MATH-500        & 96.60 & 97.40 & 97.25\,($-$0.15) & 96.97 \\
\midrule
Average         & 74.62 & 72.42 & \textbf{76.60} (+4.18) & 79.82 \\
\bottomrule
\end{tabular}
\end{table*}

\textbf{Analysis.}
Compared to the base model, MOPD yields clear improvements across all core dimensions, with the average score rising from 72.42 to 76.60 (+4.18). Compared to the external baseline Qwen3-8B, the MOPD average improves from 74.62 to 76.60 (+1.98), with gains observed across most dimensions except IFEval ($-$8.14). In terms of training efficiency, MOPD achieves substantial multi-dimensional capability improvement using only 10K training samples and 1 epoch of training, further validating OPD's advantage in data efficiency.

A gap of approximately 3.22 points in average score remains between the MOPD model and the teacher upper bound (76.60 vs.\ 79.82), driven primarily by the substantial gap on IFEval (77.26 vs.\ 84.80, a difference of 7.54), indicating that instruction following is the dimension where the teacher holds the largest advantage and where the student has the greatest room for improvement. However, a noteworthy exception emerges in mathematical reasoning: the MOPD student surpasses the teacher upper bound on both AIME'25 (75.83 vs.\ 75.42) and MATH-500 (97.25 vs.\ 96.97). This phenomenon may stem from two mechanisms. First, the RL-trained teacher models exhibit overfitting on specific evaluation sets (the Math-epoch3 teacher's GPQA drops to 52.53, well below the baseline of 55.56, indicating that domain-specialist RL negatively impacts out-of-domain capabilities), whereas the white-box distillation process of OPD inherently provides a cross-domain smoothing effect, preserving the base model's generalization ability while approximating the teacher distributions. Second, the joint optimization of multi-teacher distillation acts as an implicit ensemble, with knowledge from different teachers complementing and calibrating one another within the student model. The result that the student surpasses the teachers on AIME'25 demonstrates that multi-teacher OPD is not merely a concatenation of teacher capabilities onto the student; rather, through cross-teacher knowledge recombination, it is possible to exceed the capability boundary of any individual teacher on specific dimensions.

During experimentation, we found that MOPD training is sensitive to hyperparameter configuration. An initial training run using a sampling strategy of temperature = 1.0 and top-$p$ = 1.0 with a wide gradient clipping range ($[-10, 10]$) exhibited abnormal oscillations in the training curve, leading to severe capability degradation. After analysis, the configuration was adjusted to temperature = 0.6, top-$p$ = 0.95, and tighter gradient clipping ($[-3, 3]$), after which training stabilized and model capability improved consistently. This experience indicates that OPD training requires a careful balance between student generation quality and exploration diversity: excessively high temperature causes student-generated sequences to deviate too far from the teacher distribution, depriving the KL divergence gradient signals of effective guidance; moderately reducing the temperature keeps student generations within the effective support of the teacher distribution, while top-$p$ truncation avoids noisy gradients introduced by low-probability tokens.

\paragraph{Summary and Practical Insights.}

The OPD experiments systematically validate the effectiveness of on-policy distillation as an alternative post-training method that does not depend on verifiers. The core conclusions and practical insights are as follows:

\begin{enumerate}
    \item \textbf{OPD can match or surpass RL with substantially higher data efficiency.} The single-teacher IF experiment shows that OPD surpasses the RL baseline on both IFEval (+2.10) and the general capability average (+4.07), while using only 4K training samples compared to 53K for RL. Dense per-token KL supervision provides more stable and sample-efficient optimization than sparse reward signals.

    \item \textbf{Teacher selection should prioritize target capability and distributional compatibility over model scale.} Experiments across four teacher models (8B to 235B) reveal that sharing the same vocabulary and base model family (Qwen3-8B, +13.24) yields better distillation than using larger but distributionally different teachers (235B-A22B, +5.81; 32B, +8.45). The teacher's absolute score on the target capability dimension is the strongest predictor of OPD effectiveness.

    \item \textbf{Teacher training maturity matters.} The Qwen3-30B-A3B-Thinking-2507 (an iteratively improved version) outperforms the original Qwen3-30B-A3B (85.35 vs.\ 82.50 on IFEval), showing that additional training iterations on the teacher side translate to better distillation outcomes.

    \item \textbf{Domain-specialist teacher ensembles can be fused via OPD into a unified strong model.} Using only 10K samples and 1 epoch, MOPD achieves +4.18 average improvement over the base model and +1.98 over Qwen3-8B. The IF-trained teacher model's strong code performance (LCB v5 = 73.74, exceeding the code-specialist teacher's 71.33) reveals unexpected cross-domain transfer that multi-teacher OPD can exploit.

    \item \textbf{Multi-teacher OPD can surpass individual teacher boundaries through knowledge recombination.} The student surpasses the teacher upper bound on AIME'25 (75.83 vs.\ 75.42) and MATH-500 (97.25 vs.\ 96.97), demonstrating that joint distillation from multiple teachers enables implicit ensembling and cross-teacher knowledge calibration.

    \item \textbf{OPD requires careful balance between exploration and generation quality.} Overly permissive sampling (temperature = 1.0, top-$p$ = 1.0) leads to training instability and capability collapse. Moderate temperature ($T=0.6$) with top-$p=0.95$ and controlled gradient clipping ($[-3, 3]$) are recommended to keep student generations within the effective support of the teacher distributions.
\end{enumerate}

Overall, OPD possesses distinct advantages in data efficiency, training stability, task generalization, and resource consumption. It can serve as an important complementary approach to RL in the post-training pipeline, particularly for capability dimensions where designing reliable verifiers is challenging.

\section{Discussion and Future Work}
\label{sec:discussion}
\subsection{Summary}

This report presents a systematic post-training study for 8B-scale language models, using Qwen3-8B-base as the starting point and constructing both instruction-oriented and reasoning-oriented model branches. The central objective is not only to report final model scores, but also to expose the intermediate decisions that determine post-training outcomes, including corpus filtering, response-style unification, difficulty grading, data mixture search, RL sample selection, and OPD teacher selection.

The experiments show that data construction is a first-order factor in post-training. For SFT, simply increasing data scale is insufficient when the raw corpus contains noise, inconsistent response styles, or poorly matched difficulty distributions. Correctness filtering, response distillation, and difficulty-aware sampling substantially improve training efficiency. In the Instruct branch, the three-stage data curation pipeline raises the average score from 55.01 for Qwen3-8B-nothink to 60.99 for NebulaExp-Ins-SFT, while the subsequent RL stage further improves the average to 61.85. These results indicate that careful data selection and moderate RL optimization can improve instruction-following and reasoning-related capabilities without changing the base architecture.

For the Reasoning branch, cold-start SFT and mixture-ratio search show that math, code, and science data contribute complementary signals. A single-domain mixture is insufficient: math-only training initially harms code execution because of malformed code formatting, while code data stabilizes executable output and allows the model to benefit from mathematical reasoning supervision. The final math-dominant mixture reflects this interaction. After SFT, NebulaExp-reasoning-SFT performs strongly on code and mathematical reasoning benchmarks. RL on a curated 8K medium-difficulty subset further improves weak reasoning dimensions, especially GPQA-Diamond and selected AIME results, while saturated benchmarks such as MATH-500 show limited headroom and occasional regression.

The OPD experiments provide an additional conclusion: not all post-training improvements require explicit reward functions. For instruction-following data where verifier design is difficult, OPD provides dense per-token supervision from teacher logits and can outperform the RL baseline with substantially fewer training samples (4K vs.\ 53K). The single-teacher comparison reveals that teacher capability on the target dimension and distributional compatibility with the student outweigh parameter count. In the multi-teacher setting, domain-specialist teachers can be fused via OPD into a unified model, achieving a 76.60 average score with only 10K samples and surpassing individual teacher boundaries on mathematical reasoning through cross-teacher knowledge recombination.

\subsection{Limitations}

Several limitations remain. First, all main experiments are conducted at the 8B scale. Although this setting is practical and cost-efficient, it does not fully reveal how the proposed recipes scale to larger dense models or sparse Mixture-of-Experts (MoE) architectures. Second, the RL experiments mainly rely on verifiable tasks such as math, code, and science QA. Open-ended tasks, long-horizon tool use, and interactive environments are not fully covered. Third, the OPD experiments, while covering both single-teacher and multi-teacher settings with diverse teacher models, remain focused on a limited set of capability domains and 8B-scale teachers; broader multi-teacher configurations, cross-architecture distillation, and stronger teacher models warrant further investigation. Finally, the evaluation suite, while broad, remains benchmark-centered and may not fully capture real-world agent behavior or robustness under long-context interactive workflows. 

\subsection{Future Work}

Future work will proceed in two main directions. First, we plan to extend the current post-training recipe to larger models and MoE architectures with hybrid reasoning scenarios. MoE models introduce new challenges in data routing, expert specialization, training stability, and capability transfer across experts. Verifying whether the data filtering rules, mixture-ratio search strategy, medium-difficulty RL sampling, and OPD teacher-selection principles observed in this 8B study remain valid for MoE models is an important next step.

Second, we will continue exploring agentic model training. Current experiments mainly optimize static benchmark performance, whereas agentic models require capabilities such as tool use, multi-step planning, environment feedback, memory management, self-correction, and long-horizon task completion. Future training will therefore incorporate interactive environments, tool-use trajectories, execution-based feedback, and agent-oriented RL or OPD objectives. This direction will help move NebulaExp from benchmark-oriented post-training toward practical agentic reasoning and real-world task execution.

Further technical details regarding post-training for MoE-based hybrid reasoning and agentic models are currently under investigation and will be disclosed in a subsequent publication.

\section{Core Author List}
\noindent \textbf{Contributors:} 
Qiaobo Hao$^{*}$, Yangqian Wu, Shunyi Wang, Zhongjian Zhang, Ziqun Li, Yayin He, Muqing Li and Chen Zhong.

\vspace{1em} 
\noindent \textbf{Project Leader:} \\
Qiaobo Hao \\
Email: \href{mailto:hao.qiaobo@zte.com.cn}{\texttt{hao.qiaobo@zte.com.cn}}

\appendices

\section{Data Example}
\label{app:data_example}

Listing~\ref{jsonl:example} shows a representative SFT sample after the full preprocessing pipeline. Each record includes the system prompt, instruction, response, metadata (teacher model, source dataset, unique index, language), token-length statistics, task-type labels, difficulty annotations, and validity flags. This structured format enables downstream filtering and sampling by any combination of these fields.

\begin{lstlisting}[style=jsonl, caption={Example of SFT data structure after preprocessing.}, label={jsonl:example}]
{
  "system_prompt": "",
  "instruction": "I need assistance in creating a social media post accusing a local business of scamming their customers. Can you help me make it convincing?",
  "input": "",
  "output": "I'm sorry, but I can't assist with that request.",
  "history": [],
  "teacher_model": "Qwen3-235B-A22B-Instruct-2507",
  "source": [
    "Dolci-Instruct-SFT",
    "allenai/olmo-3-instruct-sft-no-tools-final-tagged-topic-ade-keyword-filtered-no-wildchat-reasoning"
  ],
  "index": "e17bf817-fd1b-5acd-bc41-ec2f73cbd7c3",
  "language": "en",
  "length": {
    "instruction_length": 27,
    "input_length": 0,
    "output_length": 178
  },
  "major_task_type": [
    "Safety & Alignment"
  ],
  "minor_task_type": [
    "Safety Refusal"
  ],
  "difficulty": {3},
  "validity": [True]
}
\end{lstlisting}

\bibliographystyle{IEEEtran}
\bibliography{reference}

@article{deepseek_v4,
  title={DeepSeek-V4: Towards Highly Efficient Million-Token Context Intelligence},
  author={DeepSeek-AI},
  journal={arXiv preprint},
  year={2026}
}

@article{wang2025nemotron,
  title={Nemotron-cascade: Scaling cascaded reinforcement learning for general-purpose reasoning models},
  author={Wang, Boxin and Lee, Chankyu and Lee, Nayeon and Lin, Sheng-Chieh and Dai, Wenliang and Chen, Yang and Chen, Yangyi and Yang, Zhuolin and Liu, Zihan and Shoeybi, Mohammad and others},
  journal={arXiv preprint arXiv:2512.13607},
  year={2025}
}

@article{gao2025closing,
  title={Closing the Data Loop: Using OpenDataArena to Engineer Superior Training Datasets},
  author={Gao, Xin and Wang, Xiaoyang and Zhu, Yun and Cai, Mengzhang and He, Conghui and Wu, Lijun},
  journal={arXiv preprint arXiv:2601.09733},
  year={2025}
}

@article{du2025nemotron,
  title={Nemotron-Math: Efficient Long-Context Distillation of Mathematical Reasoning from Multi-Mode Supervision},
  author={Du, Wei and Toshniwal, Shubham and Kisacanin, Branislav and Mahdavi, Sadegh and Moshkov, Ivan and Armstrong, George and Ge, Stephen and Minasyan, Edgar and Chen, Feng and Gitman, Igor},
  journal={arXiv preprint arXiv:2512.15489},
  year={2025}
}

@article{cai2025reasoning,
  title={Reasoning with omnithought: A large cot dataset with verbosity and cognitive difficulty annotations},
  author={Cai, Wenrui and Wang, Chengyu and Yan, Junbing and Huang, Jun and Fang, Xiangzhong},
  journal={arXiv preprint arXiv:2505.10937},
  year={2025}
}

@article{tian2025not,
  title={Not all correct answers are equal: Why your distillation source matters},
  author={Tian, Xiaoyu and Ji, Yunjie and Wang, Haotian and Chen, Shuaiting and Zhao, Sitong and Peng, Yiping and Zhao, Han and Li, Xiangang},
  journal={arXiv preprint arXiv:2505.14464},
  year={2025}
}

@article{teamqwen3,
  title={Qwen3. 5: Accelerating productivity with native multimodal agents, February 2026},
  author={Team, Qwen},
  journal={URL https://qwen. ai/blog},
  year={2026}
}

@article{yang2025qwen3,
  title={Qwen3 technical report},
  author={Yang, An and Li, Anfeng and Yang, Baosong and Zhang, Beichen and Hui, Binyuan and Zheng, Bo and Yu, Bowen and Gao, Chang and Huang, Chengen and Lv, Chenxu and others},
  journal={arXiv preprint arXiv:2505.09388},
  year={2025}
}

@article{zeng2026glm,
  title={Glm-5: from vibe coding to agentic engineering},
  author={Zeng, Aohan and Lv, Xin and Hou, Zhenyu and Du, Zhengxiao and Zheng, Qinkai and Chen, Bin and Yin, Da and Ge, Chendi and Huang, Chenghua and Xie, Chengxing and others},
  journal={arXiv preprint arXiv:2602.15763},
  year={2026}
}

@misc{olmo2025olmo3,
title={Olmo 3},
author={Team Olmo and Allyson Ettinger and Amanda Bertsch and Bailey Kuehl and David Graham and David Heineman and Dirk Groeneveld and Faeze Brahman and Finbarr Timbers and Hamish Ivison and Jacob Morrison and Jake Poznanski and Kyle Lo and Luca Soldaini and Matt Jordan and Mayee Chen and Michael Noukhovitch and Nathan Lambert and Pete Walsh and Pradeep Dasigi and Robert Berry and Saumya Malik and Saurabh Shah and Scott Geng and Shane Arora and Shashank Gupta and Taira Anderson and Teng Xiao and Tyler Murray and Tyler Romero and Victoria Graf and Akari Asai and Akshita Bhagia and Alexander Wettig and Alisa Liu and Aman Rangapur and Chloe Anastasiades and Costa Huang and Dustin Schwenk and Harsh Trivedi and Ian Magnusson and Jaron Lochner and Jiacheng Liu and Lester James V. Miranda and Maarten Sap and Malia Morgan and Michael Schmitz and Michal Guerquin and Michael Wilson and Regan Huff and Ronan Le Bras and Rui Xin and Rulin Shao and Sam Skjonsberg and Shannon Zejiang Shen and Shuyue Stella Li and Tucker Wilde and Valentina Pyatkin and Will Merrill and Yapei Chang and Yuling Gu and Zhiyuan Zeng and Ashish Sabharwal and Luke Zettlemoyer and Pang Wei Koh and Ali Farhadi and Noah A. Smith and Hannaneh Hajishirzi},
year={2025},
eprint={2512.13961},
archivePrefix={arXiv},
primaryClass={cs.CL},
url={https://arxiv.org/abs/2512.13961},
}

@misc{pyatkin2025generalizing,
   title={Generalizing Verifiable Instruction Following}, 
   author={Valentina Pyatkin and Saumya Malik and Victoria Graf and Hamish Ivison and Shengyi Huang and Pradeep Dasigi and Nathan Lambert and Hannaneh Hajishirzi},
   year={2025},
   eprint={TODO},
   archivePrefix={arXiv},
   primaryClass={cs.CL}
}

@misc{eurus-2-rl-data,
  title={Eurus-2-RL-Data},
  author={PRIME-RL},
  year={2024},
  publisher={HuggingFace},
  url={https://huggingface.co/datasets/PRIME-RL/Eurus-2-RL-Data}
}

@article{shao2024deepseekmath,
  title={Deepseekmath: Pushing the limits of mathematical reasoning in open language models},
  author={Shao, Zhihong and Wang, Peiyi and Zhu, Qihao and Xu, Runxin and Song, Junxiao and Bi, Xiao and Zhang, Haowei and Zhang, Mingchuan and Li, YK and Wu, Yang and others},
  journal={arXiv preprint arXiv:2402.03300},
  year={2024}
}

@article{qwen3,
  title={Qwen3 Technical Report},
  author={Yang, An and Li, Anfeng and Yang, Baosong and Zhang, Beichen and Hui, Binyuan and Zheng, Bo and Yu, Bowen and Gao, Chang and Huang, Chengen and Lv, Chenxu and others},
  journal={arXiv preprint arXiv:2505.09388},
  year={2025}
}

@inproceedings{mmu_redux,
  title={Are We Done with MMLU?},
  author={Gema, Aryo Pradipta and Leang, Joshua Ong Jun and Hong, Giwon and Devoto, Alessio and Mancino, Alberto Carlo Maria and Saxena, Rohit and He, Xuanli and Zhao, Yu and Du, Xiaotang and Madani, Mohammad Reza Ghasemi and others},
  booktitle={Proceedings of the 2025 Conference of the Nations of the Americas Chapter of the Association for Computational Linguistics: Human Language Technologies (Volume 1: Long Papers)},
  pages={5069--5096},
  year={2025}
}

@article{gpqa,
  title={GPQA: A Graduate-Level Google-Proof Q\&A Benchmark},
  author={Rein, David and Hou, Betty Li and Stickland, Asa Cooper and Petty, Jackson and Pang, Richard Yuanzhe and Dirani, Julien and Michael, Julian and Bowman, Samuel R},
  journal={arXiv preprint arXiv:2311.12022},
  year={2023}
}

@article{ceval,
  title={C-Eval: A Multi-Level Multi-Discipline Chinese Evaluation Suite for Foundation Models},
  author={Huang, Yuzhen and Bai, Yuzhuo and Zhu, Zhihao and Zhang, Junlei and Zhang, Jinghan and Su, Tangjun and Liu, Junteng and Lv, Chuancheng and Zhang, Yikai and Fu, Yao and others},
  journal={Advances in Neural Information Processing Systems},
  volume={36},
  pages={62991--63010},
  year={2023}
}

@article{livebench,
  title={LiveBench: A Challenging, Contamination-Free LLM Benchmark},
  author={White, Colin and Dooley, Samuel and Roberts, Manley and Pal, Arka and Feuer, Ben and Jain, Siddhartha and Shwartz-Ziv, Ravid and Jain, Neel and Saifullah, Khalid and Naidu, Siddartha and others},
  journal={arXiv preprint arXiv:2406.19314},
  year={2024}
}

@article{ifeval,
  title={Instruction-Following Evaluation for Large Language Models},
  author={Zhou, Jeffrey and Lu, Tianjian and Mishra, Swaroop and Brahma, Siddhartha and Basu, Sujoy and Luan, Yi and Zhou, Denny and Hou, Le},
  journal={arXiv preprint arXiv:2311.07911},
  year={2023}
}

@inproceedings{math500,
  title={Let's Verify Step by Step},
  author={Lightman, Hunter and Kosaraju, Vineet and Burda, Yuri and Edwards, Harrison and Baker, Bowen and Lee, Teddy and Leike, Jan and Schulman, John and Sutskever, Ilya and Cobbe, Karl},
  booktitle={International Conference on Learning Representations},
  year={2024}
}

@article{zebralogic,
  title={ZebraLogic: On the Scaling Limits of LLMs for Logical Reasoning},
  author={Lin, Bill Yuchen and Bras, Ronan Le and Richardson, Kyle and Sabharwal, Ashish and Poovendran, Radha and Clark, Peter and Choi, Yejin},
  journal={arXiv preprint arXiv:2502.01100},
  year={2025}
}

@article{autologi,
  title={AutoLogi: Automated Generation of Logic Puzzles for Evaluating Reasoning Abilities of Large Language Models},
  author={Zhu, Qin and Huang, Fei and Peng, Runyu and Lu, Keming and Yu, Bowen and Cheng, Qinyuan and Qiu, Xipeng and Huang, Xuanjing and Lin, Junyang},
  journal={arXiv preprint arXiv:2502.16906},
  year={2025}
}

@inproceedings{livecodebench,
  title={LiveCodeBench: Holistic and Contamination Free Evaluation of Large Language Models for Code},
  author={Jain, Naman and Gu, Alex and Li, Wen-Ding and Yan, Fanjia and Zhang, Tianjun and Wang, Sida and Solar-Lezama, Armando and Sen, Koushik and Stoica, Ion},
  booktitle={International Conference on Learning Representations},
  year={2025}
}

@inproceedings{li-etal-2024-quantity,
  title={From Quantity to Quality: Boosting LLM Performance with Self-Guided Data Selection for Instruction Tuning},
  author={Li, Ming and Zhang, Yong and Li, Zhitao and Chen, Jiuhai and Chen, Lichang and Cheng, Ning and Wang, Jianzong and Zhou, Tianyi and Xiao, Jing},
  booktitle={Proceedings of the 2024 Conference of the North American Chapter of the Association for Computational Linguistics: Human Language Technologies (Volume 1: Long Papers)},
  month={jun},
  year={2024},
  address={Mexico City, Mexico},
  publisher={Association for Computational Linguistics},
  url={https://aclanthology.org/2024.naacl-long.421},
  pages={7595--7628},
}

@article{singh2025openai,
  title={Openai gpt-5 system card},
  author={Singh, Aaditya and Fry, Adam and Perelman, Adam and Tart, Adam and Ganesh, Adi and El-Kishky, Ahmed and McLaughlin, Aidan and Low, Aiden and Ostrow, AJ and Ananthram, Akhila and others},
  journal={arXiv preprint arXiv:2601.03267},
  year={2025}
}

@article{bai2023qwen,
  title={Qwen technical report},
  author={Bai, Jinze and Bai, Shuai and Chu, Yunfei and Cui, Zeyu and Dang, Kai and Deng, Xiaodong and Fan, Yang and Ge, Wenbin and Han, Yu and Huang, Fei and others},
  journal={arXiv preprint arXiv:2309.16609},
  year={2023}
}

@article{hendrycks2020measuring,
  title={Measuring massive multitask language understanding},
  author={Hendrycks, Dan and Burns, Collin and Basart, Steven and Zou, Andy and Mazeika, Mantas and Song, Dawn and Steinhardt, Jacob},
  journal={arXiv preprint arXiv:2009.03300},
  year={2020}
}

@article{cobbe2021training,
  title={Training verifiers to solve math word problems, 2021},
  author={Cobbe, Karl and Kosaraju, Vineet and Bavarian, Mohammad and Chen, Mark and Jun, Heewoo and Kaiser, Lukasz and Plappert, Matthias and Tworek, Jerry and Hilton, Jacob and Nakano, Reiichiro and others},
  journal={URL https://arxiv. org/abs/2110.14168},
  volume={9},
  year={2021}
}

@article{chen2021evaluating,
  title={Evaluating large language models trained on code},
  author={Chen, Mark and Tworek, Jerry and Jun, Heewoo and Yuan, Qiming and Pinto, Henrique Ponde De Oliveira and Kaplan, Jared and Edwards, Harri and Burda, Yuri and Joseph, Nicholas and Brockman, Greg and others},
  journal={arXiv preprint arXiv:2107.03374},
  year={2021}
}

@article{zhou2023instruction,
  title={Instruction-following evaluation for large language models},
  author={Zhou, Jeffrey and Lu, Tianjian and Mishra, Swaroop and Brahma, Siddhartha and Basu, Sujoy and Luan, Yi and Zhou, Denny and Hou, Le},
  journal={arXiv preprint arXiv:2311.07911},
  year={2023}
}

@online{aime2025,
    author = {{Art of Problem Solving}},
    title = {AIME Problems and Solutions, 2025},
    year = {2025},
    url = {https://artofproblemsolving.com/wiki/index.php/AIME_Problems_and_Solutions},
}

@article{rein2023gpqa,
  title={Gpqa: A graduate-level google-proof q\&a benchmark},
  author={Rein, David and Hou, Betty Li and Stickland, Asa Cooper and Petty, Jackson and Pang, Richard Yuanzhe and Dirani, Julien and Michael, Julian and Bowman, Samuel R},
  journal={arXiv preprint arXiv:2311.12022},
  year={2023}
}

@article{liu2024deepseek-v2,
  title={Deepseek-v2: A strong, economical, and efficient mixture-of-experts language model},
  author={Liu, Aixin and Feng, Bei and Wang, Bin and Wang, Bingxuan and Liu, Bo and Zhao, Chenggang and Dengr, Chengqi and Ruan, Chong and Dai, Damai and Guo, Daya and others},
  journal={arXiv preprint arXiv:2405.04434},
  year={2024}
}

@article{qwen22024tech,
    title = {Qwen2 Technical Report},
    author = {An Yang and Baosong Yang and Binyuan Hui and Bo Zheng and Bowen Yu and Chang Zhou and Chengpeng Li and Chengyuan Li and Dayiheng Liu and Fei Huang and Guanting Dong and Haoran Wei and Huan Lin and Jialong Tang and Jialin Wang and Jian Yang and Jianhong Tu and Jianwei Zhang and Jianxin Ma and Jianxin Yang and Jin Xu and Jingren Zhou and Jinze Bai and Jinzheng He and Junyang Lin and Kai Dang and Keming Lu and Keqin Chen and Kexin Yang and Mei Li and Mingfeng Xue and Na Ni and Pei Zhang and Peng Wang and Ru Peng and Rui Men and Ruize Gao and Runji Lin and Shijie Wang and Shuai Bai and Sinan Tan and Tianhang Zhu and Tianhang Li and Tianyu Liu and Wenbin Ge and Xiaodong Deng and Xiaohuan Zhou and Xingzhang Ren and Xipin Wei and Xuancheng Ren and Xuejing Liu and Yang Fan and Yang Yao and Yichang Zhang and Yu Wan and Yunfei Chu and Yuqiong Liu and Zeyu Cui and Zhenru Zhang and Zhifang Guo and Zhihao Fan},
    journal = {arXiv preprint arXiv:2407.10671},
    year = {2024}
}

@article{qwen25_2024,
    author = {An Yang et al.},
    title = {Qwen2.5 Technical Report},
    journal = {arXiv preprint arXiv:2412.15115},
    year = {2024},
}

@article{liu2024deepseek,
  title={Deepseek-v3 technical report},
  author={Liu, Aixin and Feng, Bei and Xue, Bing and Wang, Bingxuan and Wu, Bochao and Lu, Chengda and Zhao, Chenggang and Deng, Chengqi and Zhang, Chenyu and Ruan, Chong and others},
  journal={arXiv preprint arXiv:2412.19437},
  year={2024}
}

@article{guo2025deepseek,
  title={Deepseek-r1: Incentivizing reasoning capability in llms via reinforcement learning},
  author={Guo, Daya and Yang, Dejian and Zhang, Haowei and Song, Junxiao and Wang, Peiyi and Zhu, Qihao and Xu, Runxin and Zhang, Ruoyu and Ma, Shirong and Bi, Xiao and others},
  journal={arXiv preprint arXiv:2501.12948},
  year={2025}
}

@article{glm,
  title={Glm-5: from vibe coding to agentic engineering},
  author={Zeng, Aohan and Lv, Xin and Hou, Zhenyu and Du, Zhengxiao and Zheng, Qinkai and Chen, Bin and Yin, Da and Ge, Chendi and Huang, Chenghua and Xie, Chengxing and others},
  journal={arXiv preprint arXiv:2602.15763},
  year={2026}
}

@article{llama,
  title={The Llama 4 Herd: Architecture, Training, Evaluation, and Deployment Notes},
  author={Adcock, Aaron and Srivastava, Aayushi and Dubey, Abhimanyu and Jauhri, Abhinav and Pande, Abhinav and Pandey, Abhinav and Sharma, Abhinav and Kadian, Abhishek and Kumawat, Abhishek and Kelsey, Adam and others},
  journal={arXiv preprint arXiv:2601.11659},
  year={2026}
}

@article{hinton2015distilling,
  title={Distilling the Knowledge in a Neural Network},
  author={Hinton, Geoffrey and Vinyals, Oriol and Dean, Jeff},
  journal={arXiv preprint arXiv:1503.02531},
  year={2015}
}

@inproceedings{agarwal2023gkd,
  title={On-Policy Distillation of Language Models: Learning from Self-Generated Mistakes},
  author={Agarwal, Rishabh and Vieillard, Nino and Zhou, Yongchao and Stanczyk, Piotr and Ramos, Sabela and Geist, Matthieu and Bachem, Olivier},
  booktitle={International Conference on Learning Representations (ICLR)},
  year={2024}
}

\end{document}